\documentclass[10pt,journal,compsoc]{IEEEtran}

\ifCLASSOPTIONcompsoc
 
  \usepackage[nocompress]{cite}
\else
  \usepackage{cite}
\fi

\ifCLASSINFOpdf
   \usepackage[pdftex]{graphicx}
\else

   \usepackage[dvips]{graphicx}

\fi
\usepackage{amsmath}
\usepackage{algorithm}
\usepackage{algorithmic}
\usepackage{array}
\usepackage[normalem]{ulem}
\useunder{\uline}{\ul}{}
\usepackage{cite}
\usepackage{amsmath,amssymb,amsfonts}
\usepackage{algorithmic}
\usepackage{graphicx}
\usepackage{textcomp}
\usepackage{xcolor}
\usepackage{subfigure}
\usepackage{lineno}
\usepackage{hyperref}
\usepackage{amsfonts,amsmath,bm}
\usepackage{mathrsfs}
\usepackage{algorithm}
\usepackage{algorithmic}
\usepackage{multirow}
\usepackage{amsmath} 
\usepackage{amsthm}
\usepackage{indentfirst}

\usepackage{fixltx2e}
\usepackage{stfloats}
\usepackage{url}
\usepackage{color}
\usepackage{bm}
\usepackage{hyperref}
\usepackage{booktabs}
\usepackage{multirow}
\usepackage{makecell}
\usepackage{amsthm,amssymb,amsfonts}
\usepackage{diagbox}
\usepackage{footnote}
\usepackage{mathrsfs}
\usepackage{amsfonts,amsmath,bm}

\newtheorem{definition}{Definition}

\begin{document}

\title{Partial Domain Adaptation without Domain Alignment}

\newtheorem{myDef}{Definition}
\newtheorem{myPro}{Problem}
\newtheorem{corollary }{Corollary }
\newtheorem{theorem}{Theorem}
\newtheorem{remark}{Remark}
\newtheorem{corollary}{\bf Corollary}

\author{Weikai~Li and Songcan~Chen
\IEEEcompsocitemizethanks{\IEEEcompsocthanksitem The authors are with College of Computer Science and Technology, Nanjing University of Aeronautics and Astronautics of (NUAA), Nanjing, 211106, China.\protect\\
E-mail: \{leeweikai; s.chen\}@nuaa.edu.cn.
\IEEEcompsocthanksitem Corresponding author is Songcan Chen.}
\thanks{Manuscript received April 19, XXXX; revised August 26, XXXX.}}

\markboth{Journal of \LaTeX\ Class Files,~Vol.~14, No.~8, August~2015}%
{Shell \MakeLowercase{\textit{et al.}}: Bare Demo of IEEEtran.cls for Computer Society Journals}
\IEEEtitleabstractindextext{%
\begin{abstract}
Unsupervised domain adaptation (UDA) aims to transfer knowledge from a well-labeled source domain to a related and unlabeled target domain with identical label space. The main workhorse in UDA is domain alignment and has proven successful. However, it is practically difficult to find an appropriate source domain with identical label space. A more practical scenario is partial domain adaptation (PDA) where the source label space subsumes the target one. Unfortunately, due to the non-identity between label spaces, it is extremely hard to obtain an ideal alignment, conversely, easier resulting in mode collapse and negative transfer. These motivate us to find a relatively simpler alternative to solve PDA. To achieve this, we first explore a theoretical analysis, which says that the target risk is bounded by both model smoothness and between-domain discrepancy. Then, we instantiate the model smoothness as an intra-domain structure preserving (IDSP) while giving up possibly riskier domain alignment. To our best knowledge, this is the first naive attempt for PDA without alignment. Finally, our empirical results on benchmarks demonstrate that IDSP is not only superior to the PDA SOTAs (e.g., $\sim$+10\% on Cl$\rightarrow$Rw and $\sim$+8\% on Ar$\rightarrow$Rw), but also complementary to domain alignment in the standard UDA.
\end{abstract}

}

\maketitle

\IEEEdisplaynontitleabstractindextext

%
\IEEEpeerreviewmaketitle


\IEEEraisesectionheading{\section{Introduction}\label{sec:introduction}}
\IEEEPARstart{C}{urrently}, unsupervised domain adaptation (UDA) has caught tremendous attention in the machine learning community, which learns an adaptive classifier for the unlabeled target domain by using a labeled source domain.
Most existing works assume that the source and target domains share an identical label set or space \cite{ben2007analysis,ben2010theory}. In this specific context, domain alignment comes to the mainstream for solving UDA, i.e., instance re-weighting \cite{sugiyama2008direct}, feature alignment \cite{pan2010domain,fernando2013unsupervised,ganin2015unsupervised} and model adaptation \cite{long2013adaptation,baktashmotlagh2013unsupervised,belkin2006manifold}. However, in practice, it tends to be extremely burdensome and quite difficult to find an ideal source domain with identical label space \cite{hoffman2018cycada}. In contrast, in the context of big data, a more practical alternative is to access a large-scale source domain while working on a relative small-scale target domain to cover the target label space, which is also known as the partial domain adaptation (PDA) \cite{cao2018partial,cao2019learning,matsuura2018twins,8951442,kim2020associative,liu2021adversarial}. Unfortunately, in such a scenario, the conventional domain alignment often fails as a result of the irrelevant source classes/categories can be mixed with target data, easier resulting in mode collapse and negative transfer \cite{matsuura2018twins}. 

To alleviate this issue, almost all existing PDA studies focus on diminishing the negative impact of irrelevant source categories between two domains by an elaborately designed reweighting approach and in turn learning the domain invariant model/representation in the shared label space \cite{cao2018partial,cao2019learning,zhang2018importance,liu2021adversarial}. Unfortunately, from an algorithmic perspective, such an additional reweighting approach can be quite risky, since exactly which irrelevant categories are unknown. Moreover, these approaches are highly rely on the pseudo labels of the target domain. However, the target samples might be wrongly predicted into the irrelevant categories in the beginning under the domain shift, as shown in Fig. \ref{fig1} (b) \cite{liu2021cycle, li2020unsupervised,johansson2019support,zhao2019learning,bouvier2020robust}, making it hard to obtain an ideal alignment. The experimental results of these methods also confirm that it is difficult to accurately identify the irrelevant categories \cite{cao2018partial,cao2019learning,zhang2018importance} in source domain. Further, even if the irrelevant source categories are correctly filtered, it is still hard to obtain a perfect alignment to guarantee that the target domain can be labeled correctly \cite{bouvier2020robust,li2020rethinking}, as shown in Fig. \ref{fig1} (c). From a theoretical perspective, several recently proposed theoretical works reveal that the domain alignment hurts the transfer-ability of representations/model \cite{bouvier2020robust,zhao2019learning}. \textcolor{black}{Consequently, these drawbacks push us to give up the domain alignment and seek an alternative for solving PDA.}

Recently, several efforts have indeed been dedicated to addressing UDA without domain alignment, even if the domains involved share the same label set. For example, Li et al, progressively anchors the target samples and in turn refine the shared subspace for knowledge transfer \cite{li2020unsupervised}. Li et al, optimizes the predictive behavior in the target domain to address UDA \cite{li2020rethinking}. Liu et al, enforces the pseudo labels to generalize across domains \cite{liu2021cycle}. As one proverb goes, "the lesser of the two evils", giving up the domain alignment is sometimes a wise choice when induced misalignment does more harm than good as validated by these works. \textcolor{black}{Nonetheless, to the best of our knowledge, so far, there has been no attempt to address PDA without domain alignment, where the misalignment is more likely occurs. At the same time, the theoretical aspect of how to transfer knowledge without domain alignment has not been well studied yet. }


\textcolor{black}{To address these issues, we derive a novel generalization error bound for domain adaptation. which combines the model smoothness and the between-domain discrepancy. Considering both the difficulty of perfect alignment and the risk of misalignment in solving PDA, we simply bypass the domain alignment and focus on the model smoothness, which guides us to optimize the target label to have a smooth structure.}
Specifically, as a proof of concept, this paper presents a quite simple PDA framework to encourage the adaptation ability of the model, which instantiates the model smoothness as a commonly-used manifold structure preserving \cite{tenenbaum2000global,belkin2006manifold,wang2018visual}. In contrast to the conventional domain alignment which indirectly learns a domain invariant representation/model for adaptation, we directly optimize the target labels to enhance the adaptation performance of the model. Doing so is also quite consistent with Vapnik's philosophy, i.e., any desired problem should be solved in a direct way \cite{vapnik2013nature}.
Furthermore, on amount of the existence of the domain shift, the manifold structure tends to vary sharply across domains. For example, several source-private samples can be located near the target samples, and some of the source samples with common labels can be far away from the target samples as shown in Fig. \ref{fig1} (e). Also, the structure information of the irrelevant categories in the source domain easily incorporates bias in learning the classifier. Thus, it can be quite perilous to leverage the structure knowledge across domains. To this end, we only consider the intra domain structure preserving (IDSP) to constrain the classifier on the target domain to alleviate the negative transfer caused by the domain gap as shown in Fig. \ref{fig1} (f). 
Notably, from both theoretical and empirical views, IDSP and domain alignment can complement each other, when ideal alignment is relatively easy to be conducted, especially in UDA. In addition, we would like to emphasize that the IDSP also can work well in solving UDA compared to the existing PDA methods. For facilitating the efforts to replicate our results, our implementation is available on GitHub \footnote{\href{https://github.com/Cavin-Lee/IDSP}{https://github.com/Cavin-Lee/IDSP}.}.
In summary, this work makes the following contributions:
\begin{figure}[t]
  \centering
  \includegraphics[width=0.5\textwidth]{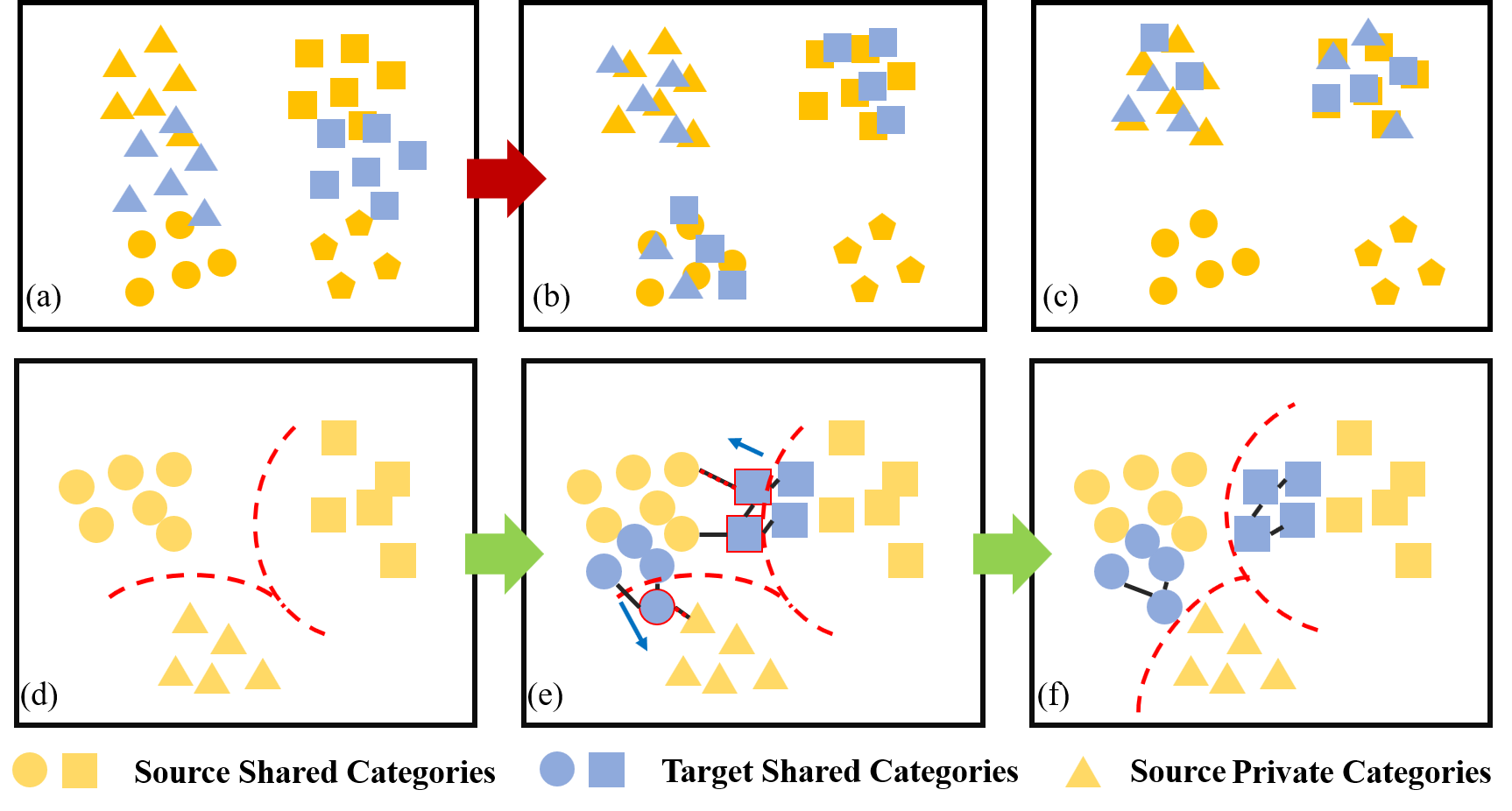}
    \caption{Motivation of IDSP. (a) Partial domain adaptation, in which the label set or space of the source domain subsumes the target one. (b) It is risky to down weight the irrelevant categories. (c) The target samples might be wrongly predicted even if the irrelevant categories are filtered. (d) Classifier learned from the source domain. (e) The manifold structure across domains tends to vary sharply due to the domain shift. (f) We only use the intra-domain structure preserving to enhance the ability of the adaptation ability of the classifier.}
  \label{fig1}
\end{figure}



\begin{itemize}
\item A novel generalization error bound is derived, which provides an alternative (i.e., model smoothness) for solving PDA;
\item A simple yet effective PDA method by intra-domain structure preserving (IDSP) is proposed, which is the first attempt, to our best knowledge, to address PDA without domain alignment;
\item The experimental results reveal that the proposed IDSP can effectively enhance the adaptation ability of the model in PDA by a significant margin on some benchmarks (e.g., $\sim$+10\% on Cl$\rightarrow$Rw and $\sim$+8\% on Ar$\rightarrow$Rw);
\item The IDSP can be complementary to domain alignment in UDA, when domain alignment is relatively less risky in such a setting, which again tells the need of targeting at a safe transfer of knowledge.
\end{itemize}

The rest of this paper is organized as follows. In Section \ref{section2}, we briefly overview UDA and PDA. In Section \ref{AA}, we elaborate on the problem formulation, and provide a novel generalization error bound for domain adaptation. In Section \ref{section3}, we present the IDSP model and its optimization algorithm in detail. In Section \ref{sec4}, we present the experimental results and the corresponding analysis. In the end, we conclude the entire paper with future research directions in Section \ref{s5}


\section{Related Works}
In this section, we present the most related research on UDA/PDA and highlight the differences between them and our method.
\label{section2}

\subsection{Unsupervised Domain Adaptation}
 Recent practices on UDA usually attempt to minimize the domain discrepancy for borrowing the existing well-established source domain knowledge. Following this, multiple domain adaptation techniques have been developed, including instance re-weighting \cite{tsuboi2009direct,sugiyama2008direct,huang2007correcting}, feature alignment \cite{pan2010domain,fernando2013unsupervised,ganin2015unsupervised} and classifier adaptation \cite{long2013adaptation,baktashmotlagh2013unsupervised,belkin2006manifold}. The instance re-weighting methods focus on correcting the distribution biases in the data sampling procedure through reweighting the individual samples to minimize the $\mathcal{A}$-distance \cite{huang2007correcting}, Maximum Mean Discrepancy (MMD) \cite{sugiyama2008direct}, or KL-divergence \cite{tsuboi2009direct}. The feature alignment methods generate the domain-invariant feature to reduce the distribution differences across domains, such as MMD \cite{pan2010domain}, Central Moment Discrepancy \cite{zellinger2017central}, Bregman divergence \cite{si2009bregman}, Joint MMD \cite{long2013adaptation,long2017deep}, $\mathcal{A}$-distance \cite{ganin2015unsupervised}, Maximum Classifier Discrepancy \cite{saito2018maximum}, Wasserstein distance \cite{courty2014domain},  $\Delta$-distance \cite{long2018conditional}, or the distance between the second-order statistics (covariance) of the source and target features \cite{sun2016return}. The classifier adaptation methods adapt the model parameter of source domain to target domain by imposing the additional constraints for alignment \cite{long2013adaptation,baktashmotlagh2013unsupervised,belkin2006manifold}. 

 The proposed IDSP can be summarized into the classifier adaptation. However, IDSP does not align the target and source domains but uses the intra-domain structure preserving to enhance the adaptation ability of the model. In addition, all of the existing UDA methods assume the source label space and target label space are identical, which is often too strict to be satisfied in the real-world applications. The proposed IDSP relaxes this assumption and aims to address a more challenging PDA problem.
\subsection{Partial Domain Adaptation}

PDA is a more practical scenario in which the source label space submerges the target one \cite{cao2018partial}. Owing to the existence of the irrelevant categories in the source domain, the risk of mis-alignment and mode collapse is highly increased. To address such issues, the existing approaches mainly focus on mitigating the potential negative transfer caused by irrelevant categories in the source domain by category-reweighting \cite{cao2018partial,cao2018partialCVPR,8951442,liu2021adversarial} or sample-reweighting \cite{matsuura2018twins,cao2019learning}.

Specifically, the category-reweighting works focus on alleviating the negative transfer caused by the source-private categories. To achieve this, Partial Adversarial Domain Adaptation (PADA) alleviates the negative transfer by down-weighting the data of irrelevant categories in the source domain with the guidance of domain discriminator \cite{cao2018partial}. Selective Adversarial Network (SAN) \cite{cao2018partialCVPR} extends PADA to maximally match the data distributions in the shared label space by adding multiple domain discriminators. Deep Residual Correction Network (DRCN) plugs one residual block into the source network to enhance the adaptation from source to target and explicitly weakens the influence from the irrelevant source categories \cite{8951442}. Conditional and Label Shift (CLS) introduce a category-level balancing parameter to align both marginal and conditional distribution between source and target domains \cite{liu2021adversarial}. 

In contrast to the mentioned category re-weighting scheme, Importance Weighted Adversarial Nets (IWAN) follows the idea of instance re-weighting to filter out the influence of the irrelevant samples in the source domain. Two Weighted Inconsistency-reduced Networks (TWINs) designs an inconsistency loss to down-weighting the outlier sample in the source domain \cite{matsuura2018twins}. Example Transfer Network (ETN) integrates the discriminative information into the sample-level weighting mechanism \cite{cao2019learning}. 

Almost all of the existing PDA methods rely on various adversarial strategies and deep networks to learn the target model with the guidance of the specifically designed category/sample reweighting approaches. However, such an approach could be vulnerable in terms of adaptation due to the agnostic label space. Besides, the weighted adversarial training based methods may suffer from the issues of training instability and mode collapse, since they are highly base on the unreliable pseudo labels \cite{li2020unsupervised}. In contrast, IDSP requires neither adversarial training nor reweighting and thus highly mitigates these issues. In addition, the IDSP does not filter out the outlier/irrelevant categories and thus can also work well on UDA.
\subsection{Model Smoothness}
\textcolor{black}{Model smoothness or model consistency aims to find a model that lies in neighborhoods having a uniformly/flat prediction output \cite{du2021efficient,foret2020sharpness}, which has been empirically and theoretically validated to improve the model generalization/robustness in the independent and identically distributed scenario  \cite{du2021efficient,foret2020sharpness,yi2021improved,miyato2018virtual}. 
In the setting of domain adaptation, several attempts are motivated to boost the DA performance by heuristically incorporating an additional model smoothness constraint into their objectives. For example, Shu et al, \cite{shu2018dirt} and Lee et al, \cite{lee2019drop} incorporate the locally-Lipschitz constraint via
virtual adversarial training to enforce classifier consistency, which enforces classifier consistency within the norm-ball neighborhood. Yang et al, \cite{yang2020phase} minimizes the negative cosine of the difference between the original and the transformed phases and increases the phase consistency of the learned model. Mishra1 et al, \cite{mishra2021surprisingly} designs a consistency regularization, which promotes the model to produce the same output for both an input image and a perturbed version. \textcolor{black}{Besides, by assuming the data lie on a manifold \cite{tenenbaum2000global}, a geometric smoothness penalty is provided to enhance the performance of domain adaptation \cite{cheng2014semi,gong2019dlow}}. Although these works achieved comparable results, they only validate the effectiveness of model smoothness in domain adaptation empirically, but fail in the theoretical analysis. In addition, existing works only treat the model smoothness/consistency as a complement in addressing domain alignment. }

\textcolor{black}{In contrast to the current studies that mainly focus on the scenario with the closed set, we focus on a more challenging task where the target label set is only a subset of the source one (i.e., PDA). In addition, we derive a brand-new target generalization bound for solving the domain adaptation, which coincidentally reveals a brand-new and complement concern, (i.e., model smoothness) rather than domain alignment. It gives birth to a model-smoothness constraint. Besides, we only utilize the model smoothness while dropping out domain alignment to address PDA since domain alignment can be risky in such setting.}

\section{A Theoretical Analysis for Domain Adaptation}\label{AA}
Current theoretical works on domain adaptation encourages domain alignment in solving domain adaptation \cite{ben2010theory,ben2007analysis}. Unfortunately, it is often hard to obtain an ideal alignment in domain adaptation, due to the large domain shifts and the un-identical label space. To find a relatively simpler alternative, we now attempt to derive a novel generalization error bound for solving the partial domain adaptation.

\subsection{Notions and Definitions}
 In this paper, we focus on the PDA and UDA scenarios. We use $\mathcal{X} \subset \mathbb{R}^d$ and $\mathcal{Y}\subset \mathbb{R}$ to denote the input and output space, respectively. The PDA and UDA scenarios constitute a labeled source domain $\mathcal{D}_s=\{x_i^s,y_i^s\}_{i=1}^n$ with $n$ samples, and an unlabeled target domain $\mathcal{D}_t=\{x_i^t\}_{i=n+1}^{n+m}$ with $m$ samples. Specifically, $x_i\in \mathcal{X}$ and $y_i \in \mathcal{Y} $ are the feature and label of $i$-th sample, respectively. The source and target domains follow the distributions $\mathbb{P}$ and $\mathbb{Q}$, respectively, and $\mathbb{P} \neq \mathbb{Q}$. Let $\mathcal{X}_s$, $\mathcal{X}_t$ and $\mathcal{Y}_s$, $\mathcal{Y}_t$ denote the feature and label spaces of source and target domains, respectively. We assume that the source and target domains share the same feature space i.e., $\mathcal{X}_s = \mathcal{X}_t$. For PDA, the target label space $\mathcal{Y}_t$ is submerged by the source label space $\mathcal{Y}_s$, i.e., $\mathcal{Y}_t \subset \mathcal{Y}_s$. Here, we further have $\mathbb{P}_s \neq \mathbb{Q}$, where $\mathbb{P}_s$ is the distribution of the share categories in the source domain. For UDA, the source and target domains share the identical label space, i.e., $\mathcal{Y}_s = \mathcal{Y}_t$. The goal of both PDA and UDA is to transfer the discriminative information from the source domain. In addition, our generalization bounds are based on Total Variation distance \cite{villani2009optimal} and model smoothness defined as follows:



\begin{definition}
\label{def2}
Total Variation distance \textnormal{\cite{villani2009optimal}}: Given two distributions $\mathbb{P}$ and $\mathbb{Q}$. The Total Variation distance $\mathrm{TV} (\mathbb{P},\mathbb{Q})$ between distributions $\mathbb{P}$ and $\mathbb{Q}$ is defined as:
\end{definition}
\begin{equation}
    \mathrm{TV}(\mathbb{P}, \mathbb{Q})=\frac{1}{2} \int_{\mathcal{X}}|d \mathbb{P}(\mathbf{x})-d \mathbb{Q}(\mathbf{x})|.
\end{equation}

\begin{definition}
\label{def3}
Model Smoothness: A model $f$ is $r$-cover with $\epsilon$ smoothness on distribution $\mathbb{P}$, if 
\end{definition}
\begin{equation}
    \mathbb{E}_{\mathbb{P}}\left[\sup _{\|\boldsymbol{\delta}\|_{\infty} \leq r}|f(\mathbf{w}, \mathbf{x}+\boldsymbol{\delta})-f(\mathbf{w}, \mathbf{x})|\right] \leq \epsilon.
\end{equation}
where $\|\boldsymbol{\cdot}\|_{\infty}$ is the $l_\infty$-norm\footnote{\textcolor{black}{In contrast to the current work which denotes the model smoothness on a Euclidean ball neighborhood, we denote it on a cube neighborhood, which contains the former one.}}.
\subsection {Generalization Error Bound with Smoothness}
Inspired by a recent theoretical study in robust learning \cite{yi2021improved}, we assume that both the source and target domains have a compact support $\mathcal{X}\in \mathbb{R}^d$. Thus, there exists $D>0$, such that $\forall {\mathbf{u,v}\in\mathcal{X}},\Vert \mathbf{u-v} \Vert <D$. Let $\mathcal{L}(f(\mathbf{x}),y)$ be the continuous and differentiable loss function.  We represent the expected risk of model $f$ over distribution $\mathbb{P}$ as  $\mathcal{E}_{\mathbb{P}}\left(f\right) = \mathbb{E}_{\{x,y\}\sim \mathbb{P}}\mathcal{L}(f(x),y)$. In addition, we also assume that $0\leq \mathcal{L}(f(\mathbf{x}),y) \leq M$ for constant $M$ without loss of generality.
\begin{theorem}
\label{the2}
 Given two distributions $\mathbb{P}$ and $\mathbb{Q}$, if a model $f$ is $2r$-cover with $\epsilon$ smoothness over distributions $\mathbb{P}$ and $\mathbb{Q}$, with probability at least $ 1 - \theta$, we have:
\end{theorem}
\begin{equation}
\label{eq13}
\begin{aligned}
   \mathcal{E}_{\mathbb{Q}}\left(f\right) 
   \leq  \mathcal{E}_{\mathbb{P}}\left(f\right) +2&\epsilon+ 2M\mathrm{TV}(\mathbb{P},\mathbb{Q})\\
    +  M&\sqrt{\frac{\left(2 d\right)^{\frac{2 \epsilon^{2} D}{r^{2}}+1} \log 2+2 \log \left(\frac{1}{\theta}\right)}{m}}\\
      +  M&\sqrt{\frac{\left(2 d\right)^{\frac{2 \epsilon^{2} D}{r^{2}}+1} \log 2+2 \log \left(\frac{1}{\theta}\right)}{n}}\\
       +  M&\sqrt{\frac{\log(1/\theta)}{2m}}.
\end{aligned}
\end{equation} 

The proof is given in the Appendix A. \textcolor{black}{Note that both $M$ and $D$ are some constants for the given dataset, this generalization error bound ensures that the target risk $\mathcal{E}_{\mathbb{Q}}\left(f\right)$ is bounded by the source risk $\mathcal{E}_{\mathbb{P}}\left(f\right)$, the model smoothness $\epsilon$ and the domain discrepancy $\mathrm{TV}(\mathbb{P},\mathbb{Q})$.} While the misalignment tends to be risky for solving PDA, we have to give up domain alignment and turn to focus on decreasing the $\epsilon$ to guarantee a low target risk. In other words, if two points $x_1, x_2 \in \mathcal{X} $ are close, then the  $f(x_1)$ and $f(x_2)$ should be similar.

\section{Intra Domain Structure Preserving} \label{section3}
In this section, we exploit the theoretical results introduced above to derive a simple and practical PDA algorithm for proof of concept. Specifically, we focus on model smoothness and propose an Intra Domain Structure Preserving (IDSP) regularizer to address PDA. In the following, we go through the details of IDSP. 
\subsection{Main Idea}
Our goal is to learn an adaptive classifier for the target domain $\mathcal{D}_t$. To begin with, we suppose that the classifier be $f=\mathbf{w} ^\intercal \phi(\mathbf{x})$, where $\phi(\cdot)$ denotes the feature mapping function that projects the original feature space to the Hilbert space $\mathcal{H}_K$ induced by kernel function $K(\cdot , \cdot)$, and $\mathbf{w}$ is the parameter of the classifier. Then, the learning framework is formulated as:
\begin{equation}\label{eq1}
    f = \mathop{\arg\min}_{f\in \mathcal{H}_K}\mathcal{L}(f(x),y)+\lambda \mathcal{R}(f)+\gamma \mathcal{M}(f),
\end{equation}
where the first term refers to the loss function on data samples, which represents the empirical risk of the training data or the source domain. The second term refers to the regularization term of structural risk minimization (SRM) \cite{vapnik1999overview}. The third term refers to the regularization term of model adaptation and knowledge transfer. $\lambda$ and $\gamma$ are regularization hyper-parameters accordingly. The main idea of this paper is to incorporate an appropriate and theoretically derived regularization term $\mathcal{M}(f)$ to learn a classifier $f$.

Specifically, we attempt to give up the domain alignment, since misalignment tends to do more harm than good in PDA \cite{cao2018partial} and even in UDA \cite{bouvier2020robust}. Motivated by the theoretical analysis, we turn to focus on the model smoothness. As a proof of concept, we assume that the data lies in a manifold and follow a study \cite{belkin2006manifold} and formulate $\mathcal{M}(f)$ as Intra Domain Structure Preserving \footnote{To realize the model smoothness, we also can incorporate other consistency regularization terms, such as implicit data disturbance (e.g., Jocabian norm ) and explicit data disturbance (e.g., robust constraint) by assuming the data lies in the Euclidean space \cite{chen2020simple, he2020momentum}. More details are listed in the Appendix D: }.

\subsection{Structural Risk Minimization}
To learn the classifier $f$, we only calculate the empirical risk on the source domain, since the target domain has no label information. Then, we minimize the structural risk functional as:
\begin{equation}\label{eq2}
    f=\underset{f \in \mathcal{H}_{K}}{\arg \min } \sum_{i=1}^{n}\left(y_{i}-f\left(\mathbf{x}_{i}\right)\right)^{2}+\lambda\|f\|_{K}^{2},
\end{equation}
where $\|f\|_{K}^2$ is the squared norm of $f$ in $\mathcal{H}_{K}$ \textcolor{black}{and $y_i\in\mathbb{R}^C$ is the one hot label vector, such that $y(c)=1$ if $x$ belongs to the $c$-th category and $y(c)=0$ otherwise. Notably, the theoretically derived model-smoothness can methodologically be adapted to almost all learning paradigms, such as DNN, deep forest, SVM, SOM, etc. Undoubtedly, different learning paradigm will likely affect the final results differently, in terms of “No Free Lunch” theorem. Thus, we select relatively simpler and steerable while more theoretically-grounded kernel regression as a basic method to avoid some confounding effects compared to complex method. In addition, since the loss function has a significant influence on the final performance, we adopt the most commonly-used square loss \cite{rifkin2003regularized} as the loss functions like MEDA \cite{wang2018visual}}. 
\subsection{Intra Domain Structure Preserving}
For a safe transfer of knowledge, we only focus on model smoothness. \textcolor{black}{Specifically, by assuming that the data lies on the manifold, the model smoothness term will naturally relax to a geometric smoothness penalty \cite{niyogi2013manifold}}. Thus, we model it as the intra domain structure preserving. Specifically, as the manifold structure may vary sharply cross domains due to the domain shift and the structure information of the irrelevant categories in the source domain easily introduce bias in the learned classifier, it tends to be risky to leverage the structure information of the source domain. Thus, we only add a Laplacian regularization term on target domain to preserve its intra domain structure. In the following, we present the graph construction and Laplacian regularization of IDSP.

\subsubsection{Graph Construction}
 The core for structure preserving is the graph construction. The pair-wise affinity matrix $\hat{\mathbf{G}}$ of the graph can be formulated as follows:

\begin{equation}
  \begin{aligned}
\hat{G}_{i j} & = \left\{\begin{array}{ll}
\operatorname{sim}\left(\mathbf{x}_{i}, \mathbf{x}_{j}\right), & \mathbf{x}_{i} \in \mathcal{N}_{p}\left(\mathbf{x}_{j}\right) \\
0, & \text { otherwise }
\end{array}\right.
\end{aligned},
\end{equation}
 where $\operatorname{sim}(\cdot, \cdot)$ denotes a proper similarity measurement (this paper use the cosine distance) between two samples. $\mathcal{N}_{p}$ represents the set of $p$-nearest neighbors of point $\mathbf{x}_{i}$. Since we only consider the intra domain structure preserving of target domain, we further have the pair-wise affinity matrix $\mathbf{G}$:
\begin{equation}
  \begin{aligned}
  \label{eq7}
G_{i j} & = \left\{\begin{array}{ll}
\hat{G}_{i j}, & \mathbf{x}_{i} \text { and } \mathbf{x}_{j} \in \mathcal{D}_{t}  \\
0, & \text { otherwise }
\end{array}\right.
\end{aligned}.
\end{equation}
Notably, if we further have $G_{i j}=\hat{G}_{i j}$ where $\mathbf{x}_{i} \text { and } \mathbf{x}_{j} \in \mathcal{D}_{s}$, we will preserve the intra structure of both source and target domain (ST). In addition, if $\forall{i,j}$, $G_{i j}=\hat{G}_{i j}$, the model will degenerate to the conventional manifold structure preserving (CST). 
\subsubsection{Laplacian Regularization}
 By introducing Laplacian matrix $\mathbf{L=D-G}$, where $D_{ii} = \sum_{i=1}^{n+m}{G_{ij}}$ is the diagnoal matrix, the Laplacian regularization for  $\mathcal{M}(f)$ is expressed by
\begin{equation} \label{eq5}
    \begin{aligned}
\mathcal{M}(f) &=\sum_{i, j=1}^{n+m}\left(f\left(\mathbf{x}_{i}\right)-f\left(\mathbf{x}_{j}\right)\right)^{2} G_{i j} \\
&=\sum_{i, j=1}^{n+m} f\left(\mathbf{x}_{i}\right) L_{i j} f\left(\mathbf{x}_{j}\right).\\
\end{aligned}
\end{equation}
\subsection{Overall Reformulation}

Substituting with Eqs.  \ref{eq2} and \ref{eq5} in Eq. \ref{eq1}, the overall reformulation is reformulated as:

\begin{equation}\label{eq6}
\begin{aligned}
 f=\underset{f \in \mathcal{H}_{K}}{\arg \min } \sum_{i=1}^{n}&\left(y_{i}-f\left(\mathbf{x}_{i}\right)\right)^{2}+\lambda\|f\|_{K}^{2}\\
 +&\gamma \sum_{i, j=1}^{n+m} f\left(\mathbf{x}_{i}\right) L_{i j}f\left(\mathbf{x}_{j}\right),
\end{aligned}
\end{equation}

\noindent
{\textbf{Remark:} We need to state that our structure preserving is based on the manifold assumption, other constraint, e.g., Jacobian Norm \cite{li2022jacobian} regularization  or other consistency regularization \cite{shu2018dirt} can be adopted to optimize the model smoothness. More details  are given in the Appendix. }
\subsection{Learning Algorithm}
The major difficulty of the optimization lies in that the kernel mapping $\phi : \mathcal{X}\rightarrow \mathcal{H}_K$ may have infinite dimensions. To solve Eq. \ref{eq6} effectively, we reformulate it by using the following revised Representer theorem.
\begin{theorem}
\label{the1}
 Representer Theorem: The parameter $\mathbf{W}^{*}=\left[\mathbf{w}_{1}^{*}, \cdots, \mathbf{w}_{h}^{*}\right]$ for the optimized solution $f$ of Eq. \ref{eq6} can be expressed in terms of the cross-domain labeled and unlabeled examples, 
\end{theorem}
\begin{equation}
\label{eqtt}
    f(\mathbf{x})=\sum_{i=1}^{n+m} \alpha_{i} K\left(\mathbf{x}_{i}, \mathbf{x}\right) \quad \text { and } \quad \mathbf{w}=\sum_{i=1}^{n+m} \alpha_{i} \phi\left(\mathbf{x}_{i}\right),
\end{equation}
where $K$ is a kernel induced by $\phi$, $\alpha_i$ is a weighting coefficient. The proof is given in the Appendix B.

By incorporating Eq. \ref{eq7} into Eq. \ref{eq6}, we obtain the following objective \footnote{\textcolor{black}{The proposed model smoothness or IDSP regularizer can be seamlessly applied to almost all learning paradigms by reformulating the $f(x)$  to the corresponding objective if once need. The reason for choosing the kernel methods is three folds. (1) the kernel methods still flourish in the machine learning community, which is still appearing in the current top journals or conferences (typically,   \cite{gizewski2022regularization,elesedy2021provably}). (2) The current DA works based on model output mostly utilize the kernel regression \cite{wang2018visual}. (3) The kernel method can be relevant to better understanding deep learning methods, since a deep network can be equivalent to a kernel under some conditions  \cite{zhang2020kernel,lee2022neural,wang2021deepdd,castellanosscalable}. }}:

\begin{equation}
    \begin{aligned}
\alpha=& \underset{\alpha \in \mathbb{R}^{n+m}}{\arg \min }\left\|\left(\mathbf{Y}-\boldsymbol{\alpha}^{\mathrm{T}} \mathbf{K}\right) \mathbf{V}\right\|_{F}^{2} \\
&+\operatorname{tr}\left(\lambda \boldsymbol{\alpha}^{\mathrm{T}} \mathbf{K} \boldsymbol{\alpha}+\gamma\boldsymbol{\alpha}^{\mathrm{T}} \mathbf{K}\mathbf{L} \mathbf{K} \boldsymbol{\alpha}\right) .
\end{aligned}
\end{equation}
where $\mathbf{V}$ is the indicator matrix with $\mathbf{V}_{ii}=1$ if $i \in \mathcal{D}_s$, otherwise$\mathbf{V}_{ii}=0$.
Setting derivative of objective function as 0 leads to:

\begin{equation}
\label{eq12}
    \boldsymbol{\alpha}=((\mathbf{V}+\gamma \mathbf{L}) \mathbf{K}+\lambda \mathbf{I})^{-1} \mathbf{V} \mathbf{Y}^{\mathbf{T}} .
\end{equation}
The learning algorithm is summarized in Algorithm \ref{alg1}.

\begin{algorithm}[ht]
    \caption{Learning algorithm for IDSP}
    \label{alg1} 
    \begin{algorithmic}[1]
    \REQUIRE ~~\\
    $n$ source labeled datasets $\mathcal{D}_s=\{\mathbf{x}^s_i,y_i^s\}_{i=1}^n$\\
    $m$ target unlabeled datasets $\mathcal{D}_t=\{\mathbf{x}^t_i\}_{i=n+1}^{m+n}$\\
    Hyper-parameters $\lambda$, $\gamma$,$p$;\\ 

    \ENSURE ~~\\ 
    Predictive Classifier $f$
    \STATE Calculate the graph Laplacian $\mathbf{L}$ by Eq.(\ref{eq7})
    \STATE Construct kernel $\mathbf{K}$  by a specific kernel function;
    \STATE Compute $\mathbf{\alpha}$ by Eq.(\ref{eq12});
    \STATE Return Classifier $f$  by Eq.(\ref{eqtt});
    \end{algorithmic}
\end{algorithm}

\subsection{Complexity Analysis}
 The computational complexity for solving IDSP consists of three parts. We denote $s$ as the average number of non-zero features of each sample and we have $s \leq d, p \ll \min (n+m, d)$.  Solving the Eq. \ref{eq12} by LU decomposition requires $\mathcal{O}((n+m)^3)$. For constructing the graph Laplacian matrix $\mathbf{L}$, IDSP requires $\mathcal{O}(sm^2)$. For constructing the kernel matrix  $\mathbf{L}$, IDSP requires $\mathcal{O}((n+m)^2)$. Thus, the total complexity of IDSP is  $\mathcal{O}((n+m)^3+sm^2+(n+m)^2)$.
\textcolor{black}{It should be noted that it is not difficult to speed up the algorithm using randomized sketching kernel learning \cite{yin2022distributed}, sample condensation  \cite{winter2021layer}, Divide-and-Conquer  \cite{yin2020divide}, conjugate gradient  \cite{long2013adaptation} or kernel approximation methods \cite{rahimi2007random}, which is beyond the scope of this work.}


\section{Experiments}
\label{sec4}
To evaluate the performance of IDSP, we conduct multiple experiments over both the PDA and the UDA settings on the most widely-used benchmark datasets including \emph{Office-Home}, \emph{Image-Clef} and \emph{Office-31}. Table \ref{table1} lists the statistics of the three datasets.

\begin{table}[]

\centering
\caption{Statistics of the benchmark datasets}
\begin{tabular}{cccc}
\hline\hline
Dataset   & \#Sample & \#Class & \#Domain \\ \hline\hline
Office-Home & 15500  & 65   & Ar, Cl, Pr, Rw    \\ 
Image-Clef & 7200   & 12   & C, I, P    \\ 
Office-31  & 4652   & 31   & A, W, D    \\ \hline\hline
\end{tabular}

\label{table1}
\end{table}

\subsection{IDSP on PDA}
\subsubsection{Datasets for PDA}
We first evaluate the performance of IDSP on the widely-adopted PDA benchmarks, i.e., \emph{Office-31} and \emph{Office-Home}. The details of the datasets are given as follows:

\begin{table*}[]

\setlength{\parindent}{2em}
  \centering
  \caption{Accuracy (\%) on \emph{Office-Home} for PDA from 61 categories to 25 categories}
  \begin{tabular}{cccccc|cccccc}
\hline\hline
      &\multicolumn{5}{c|}{UDA Methods}  &\multicolumn{6}{c}{PDA Methods}\\  
      & ResNet & DAN  & DANN & MEDA & PAS & PADA & DRCN & IWAN & SAN  & ETN  & IDSP \\\hline\hline
Ar→Cl & 38.6   & 44.4 & 44.9 & 52.1 & 55.0 & 52.0 & 54.0 & 53.9 & 44.4 & {\ul 59.2} & \textbf{60.8} \\
Ar→Pr & 60.8   & 61.8 & 54.1 & 73.8 & 75.7 & 67.0 & 76.4 & 54.5 & 68.7 & {\ul77.0} & \textbf{80.8} \\
Ar→Rw & 75.2   & 74.5 & 69.0 & 78.9 & {\ul 83.9} & 78.7 & 83.0 & 78.1 & 74.6 & 79.5 & \textbf{87.3} \\
Cl→Ar & 39.9   & 41.8 & 36.3 & 57.9 & 66.3 & 52.2 & 62.1 & 61.3 & {\ul67.5} & 62.9 & \textbf{69.3} \\
Cl→Pr & 48.1   & 45.2 & 34.3 & 61.8 & 73.3 & 53.8 & 64.5 & 48.0 & 65.0 & {\ul65.7} & \textbf{76.0} \\
Cl→Rw & 52.9   & 54.1 & 45.2 & 71.1 & 74.7 & 59.0 & 71.0 & 63.3 & {\ul77.8} & 75.0 & \textbf{80.2} \\
Pr→Ar & 49.7   & 46.9 & 44.1 & 59.7 & 65.7 & 52.6 & 70.8 & 54.2 & 59.8 & {\ul68.3} & \textbf{74.7} \\
Pr→Cl & 30.9   & 38.1 & 38.0 & 48.5 & {\ul 55.5} & 43.2 & 49.8 & 52.0 & 44.7 & 55.4 & \textbf{59.2} \\
Pr→Rw & 70.8   & 68.4 & 68.7 & 77.6 & 79.4 & 78.8 & 80.5 & 81.3 & 80.1 & {\ul84.4} & \textbf{85.3}\\
Rw→Ar & 65.4   & 64.4 & 53.0 & 68.6 & 71.8 & 73.7 & {\ul77.5} & 76.5 & 72.2 & 75.7 & \textbf{77.8} \\
Rw→Cl & 41.8   & 45.4 & 34.7 & 53.0 & 54.3 & 56.6 & {\ul59.1} & 56.7 & 50.2 & 57.7 & \textbf{61.3} \\
Rw→Pr & 70.4   & 68.8 & 46.5 & 78.5 & 82.0 & 77.1 & 79.9 & 82.9 & 78.7 & {\ul85.5} & \textbf{85.7} \\
AVE   & 53.7   & 54.5 & 47.4 & 65.1 & 69.8 & 62.1 & 69.0.& 63.6 & 65.3 & {\ul70.5} & \textbf{74.9} \\ \hline\hline
  \end{tabular}
  
  \label{table2}
  \footnotesize{ All PDA methods are achieved by the weighted domain alignment approach except IDSP, which is the same with the following Table \ref{table3}. Both MEDA and IDSP incorporate the manifold regularization term. Both PAS and IDSP are the non-aligned methods. The best accuracy is presented in bold and the second best is underlined, similarly hereinafter. }\\
\end{table*}

\begin{table*}[]

\setlength{\parindent}{2em}
  \centering
  \caption{Accuracy (\%) on \emph{Office-31} for PDA from 31 categories to 10 categories}
  \begin{tabular}{cccccc|cccccccc}
\hline\hline
    &\multicolumn{5}{c|}{UDA Methods}  &\multicolumn{8}{c}{PDA Methods}\\
    & ResNet & DAN  & DANN & MEDA & PAS  & PADA & TWINs & DRCN & IWAN & SAN  & ETN & CLS  & IDSP \\ \hline\hline
A→W & 54.5   & 46.4 & 41.4 & 79.3 & 97.0 & 86.5 & 86.0  & 90.8 & 89.2 & 93.9 & 94.5 & {\ul99.6} & \textbf{99.7} \\ 
D→W & 94.6   & 53.6 & 46.8 & 97.0 & 99.3 & 99.3 & 99.3  & \textbf{100}  & 99.3 & 99.3 & \textbf{100} & \textbf{100} & {\ul99.7} \\ 
W→D & 94.3   & 58.6 & 38.9 & {\ul99.4} &\textbf{100} & \textbf{100}  & \textbf{100}   & \textbf{100}  & {\ul99.4} & \textbf{100}& \textbf{100}  & \textbf{100}  & \textbf{100}  \\ 
A→D & 65.6   & 42.7 & 41.4 & 85.3 & 98.4 & 82.2 & 86.8  & 86.0 & 90.5 & 82.2 & 95.0 &{\ul97.3}& \textbf{99.4} \\ 
D→A & 73.2   & 65.7 & 41.3 & 92.0 & 94.6 & 92.7 & 94.7  & 95.6 & 95.6 & 92.7 & {\ul 96.2}&\textbf{97.9} & 95.1 \\ 
W→A & 71.7   & 65.3 & 44.7 & 91.6 & 94.4 & 95.4 & 94.5  & 95.8 & 94.3 & 95.4 & 94.6 & \textbf{98.3}&{\ul 95.7} \\ 
AVE & 75.6   & 55.4 & 42.4 & 90.7 & 97.2& 92.7 & 93.6  & 94.3 & 94.7 & 92.7 & 96.7 & {\ul98.2} & \textbf{98.3} \\ \hline\hline
  \end{tabular}
  
  \label{table3}
\end{table*}
\noindent
\textbf{Office-31} \cite{saenko2010adapting} contains 4652 images with 31 categories in three visual domains including Amazon(A), DSLR(D) and Webcam(W). For the setting of PDA, we follow the same splits used in recent PDA studies \cite{cao2018partial,cao2018partialCVPR} and the target domains only contain 10 categories, which is shared by \emph{Office-31} and  \emph{Caltech-256}.

\noindent
\textbf{Office-Home} is released at CVPR’17 \cite{venkateswara2017deep}, which contains 65 different objects from 4 domains including 15588 images: Artistic images (Ar), Clipart images (Cl), Product images (Pr) and Real-world images (Rw). Reference to the PDA setting in recent studies \cite{cao2018partial,cao2018partialCVPR}, the first 25 categories (in alphabetical order) are taken as the target categories, while the others as the source private categories.\\

\subsubsection{Experimental Setup}
In order to evaluate the performance of IDSP on the PDA setting, we compare IDSP with several PDA SOTAs:  Partial Adversarial Domain Adaptation (PADA) \cite{cao2018partial}, Selective Adversarial Network (SAN) \cite{cao2018partialCVPR}, Two Weighted Inconsistency-reduced Networks (TWINs) \cite{matsuura2018twins}, Importance Weighted Adversarial Network (IWAN) \cite{zhang2018importance}, Example Transfer Network (ETN) \cite{cao2019learning}, Deep Residual Correction Network (DRCN) \cite{8951442} and Conditional and Label Shift (CLS) \cite{liu2021adversarial}. To illustrate the difficulty of domain alignment in solving PDA, we further compare several traditional learning and UDA benchmarks including ResNet \cite{he2016deep}, Manifold Embedded Distribution Alignment (MEDA) \cite{wang2018visual}, Deep Adaptation Network (DAN) \cite{long2015learning} and Domain Adversarial Neural Networks (DANN) \cite{ganin2015unsupervised}. For the non-aligned method, we only compared Progressive Adaptation of Subspaces (PAS) \cite{li2020unsupervised}, since only PAS performed the PDA experiment in the original paper.  For a fair comparison, we use the 2048-dimensional deep feature (extracted using ResNet50 pre-trained on ImageNet) for both IDSP and other shallow UDA approach (i.e., MEDA and PAS). The optimal parameters of all compared methods are set following their original papers. Note that several results are directly obtained from the published papers if we follow the same setting. As for IDSP, we empirically set the hyper-parameters $\lambda=0.1$, $\gamma=5$ and $p=10$ for the PDA setting. To evaluate the performance, we follow the widely used \textbf{accuracy} as a measurement. 


\subsubsection{Experimental Results}
The classification results of 12 PDA tasks on \emph{Office-Home} dataset and 6 PDA tasks on \emph{Office-31} dataset are given in Table \ref{table2} and Table \ref{table3}, respectively. Specifically, on both two datasets, our approach achieves the superior results with average accuracies of 74.9\% on \emph{Office-31} dataset and 98.3\% on \emph{Office-31} dataset by using a simple laplacian regularization term. IDSP outperforms the state of the art
by a significant margin on several task (~+10\% on Cl$\rightarrow$Rw, ~+8\% on Ar$\rightarrow$Rw and +6\% on Cl$\rightarrow$Ar). In addition, IDSP achieves the best results on all 12 tasks at \emph{Office-Home} dataset and achieves the best/second best on all 6 tasks at \emph{Office-31} dataset. Notably, the UDA methods (e.g., DAN, DANN) work even worse than the baseline method without DA (i.e., ResNet), the reason is that the risk of mode collapse and mis-alignment is highly increased in the PDA setting. Moreover, it should be noted that MEDA consists of both structure preserving and domain alignment, whose results have a huge gap between IDSP (i.e., $\sim$ 10\% on \emph{Office-Home} and $\sim$ 8\% on \emph{Office-31}), which illustrates that the domain alignment tends to hurt the adaptation and causes the negative transfer in the PDA setting. Also, MEDA achieves better results than PADA in \emph{Office-Home} dataset, which also validates the performance gain brought by the model smoothness. In addition, PAS also achieves a quite comparable results on PDA, which reveals that giving up domain alignment is a wise choice in PDA. To sum up, the results reveal the effectiveness of the IDSP for solving PDA, while perfect alignment is quite hard to obtain in PDA setting.

\begin{table*}[]
\caption{Accuracy (\%) on \emph{Office-Home} for UDA with closed-set}
  \centering
  \begin{tabular}{ccccccccccccccc}\hline\hline
      & ResNet & 1NN  & TCA  & TJM  & CORAL & GFK  & SA   & DAN  & DANN & JAN  & CDAN &PAS & MEDA & IDSP \\\hline\hline
Ar→Cl & 34.9   & 45.3 & 38.3 & 38.1 & 42.2  & 38.9 & 43.6 & 43.6 & 45.6 & 45.9 & 46.6 &52.2& \textbf{55.2} & {\ul55.0} \\
Ar→Pr & 50.0   & 57.0 & 58.7 & 58.4 & 59.1  & 57.1 & 63.3 & 57.0 & 59.3 & 61.2 & 65.9 &72.9& \textbf{76.2} & {\ul74.5} \\
Ar→Rw & 58.0   & 45.7 & 61.7 & 62.0 & 64.9  & 60.1 & 68.0 & 67.9 & 70.1 & 68.9 & 73.4 &{\ul76.9}& \textbf{77.3} & 76.3 \\
Cl→Ar & 37.4   & 57.0 & 39.3 & 38.4 & 46.4  & 38.7 & 47.7 & 45.8 & 47.0 & 50.4 & 55.7 &{\ul58.4}& 58.0 & \textbf{59.1} \\
Cl→Pr & 41.9   & 58.7 & 52.4 & 52.9 & 56.3  & 53.1 & 60.7 & 56.5 & 58.5 & 59.7 & 62.7 &68.1& \textbf{73.7} & {\ul71.0} \\
Cl→Rw & 46.2   & 48.1 & 56.0 & 55.5 & 58.3  & 55.5 & 61.9 & 60.4 & 60.9 & 61.0 & 64.2 &69.7& \textbf{71.9} & {\ul70.4} \\
Pr→Ar & 38.5   & 42.9 & 42.6 & 41.5 & 45.4  & 42.2 & 48.2 & 44.0 & 46.1 & 45.8 & 51.8 &58.3 & {\ul59.3} & \textbf{60.4} \\
Pr→Cl & 31.2   & 42.9 & 37.5 & 37.8 & 41.2  & 37.6 & 41.5 & 43.6 & 43.7 & 43.4 & 49.1 &47.4& {\ul52.4} & \textbf{52.7} \\
Pr→Rw & 60.4   & 68.9 & 64.1 & 65.0 & 68.5  & 64.6 & 70.0 & 67.7 & 68.5 & 70.3 & 74.5 &76.6& \textbf{77.9} & {\ul77.6} \\
Rw→Ar & 53.9   & 60.8 & 52.6 & 53.0 & 60.1  & 53.7 & 59.4 & 63.1 & 63.2 & 63.9 & 68.2 &67.1& 68.2 & \textbf{68.9} \\
Rw→Cl & 41.2   & 48.3 & 41.7 & 42.0 & 48.2  & 42.3 & 47.4 & 51.5 & 51.8 & 52.4 & 56.9 &53.5& \textbf{57.5} & {\ul56.5} \\
Rw→Pr & 59.9   & 74.7 & 70.5 & 71.4 & 73.1  & 70.6 & 74.6 & 74.3 & 76.8 & 76.8 & 80.7 &77.6& {\ul81.8} & \textbf{82.1} \\
AVE   & 46.1   & 56.4 & 51.3 & 51.3 & 55.3  & 51.2 & 57.2 & 56.3 & 57.6 & 58.3 & 62.8 &64.9& \textbf{67.5} & {\ul67.0}\\ \hline\hline
\end{tabular}

  \label{tab4}
\end{table*}

\begin{table*}[htbp]
\caption{Accuracy (\%) on \emph{Image-Clef} for UDA with closed-set}
\centering
\begin{tabular}{ccccccccccccccc}\hline\hline
    & ResNet & 1NN  & TCA  & TJM  & CORAL & GFK  & SA   & DAN  & DANN & JAN  & CDAN & PAS & MEDA & IDSP \\\hline\hline
C→I & 78.0   & 83.5 & 89.3 & 90.0 & 83.0  & 86.3 & 88.2 & 86.3 & 87.0 & 89.5 & 91.3 & 90.5 & \textbf{92.7} & {\ul91.2} \\
C→P & 65.5   & 71.3 & 74.5 & 75.0 & 71.5  & 73.3 & 74.3 & 69.2 & 74.3 & 74.2 & 74.2 & {\ul75.5} & \textbf{79.1} & 74.7 \\
I→C & 91.5   & 89.0 & 93.2 & 94.2 & 88.7  & 93.0 & 94.5 & 92.8 & 96.2 & 94.7 & \textbf{97.7} & 95.1 & {\ul96.2} & 95.7 \\
I→P & 74.8   & 74.8 & 77.5 & 76.2 & 73.7  & 75.5 & 76.8 & 74.5 & 75.0 & 76.8 & 77.7 & 78.3 & \textbf{80.2} & {\ul78.5} \\
P→C & 91.2   & 76.2 & 83.7 & 85.3 & 72.0  & 82.3 & 93.5 & 89.8 & 91.5 & 91.7 & 94.3 & 95.5 & \textbf{95.8} & {\ul95.7} \\
P→I & 83.9   & 74.0 & 80.8 & 80.3 & 71.3  & 78.0 & 88.3 & 82.2 & 86.0 & 88.0 & 90.7 & 92.0 & \textbf{91.5} & \textbf{91.5} \\
AVE & 80.7   & 78.1 & 83.2 & 83.5 & 76.7  & 81.4 & 85.9 & 82.5 & 85.0 & 85.8 & 87.7 & 87.8 & \textbf{89.3} & {\ul87.9}\\ \hline\hline
\end{tabular}
\label{tab5}

\end{table*}

\subsection{IDSP on UDA }
\subsubsection{Dataset}
We also validate the performance of IDSP on the UDA setting. Two datasets including \emph{Office-Home}  and \emph{Image-Clef} are adopted, which are both the commonly-used benchmark datasets for the closed-set UDA and widely adopted in the most existing works such as \cite{pan2010domain,ganin2015unsupervised,long2017deep,long2015learning,venkateswara2017deep}. We select all categories in the \emph{Office-Home} for the UDA setting. The statistic information of the \emph{Image-Clef} is given as follows.\\

\noindent
\textbf{Image-Clef} \cite{long2017deep} derives from Image-Clef 2014 domain adaptation challenge, and is organized by selecting 12 object categories shared in the three famous real-world datasets, ImageNet ILSVRC 2012 (I), Pascal VOC 2012 (P), Caltech-256 (C). It includes 50 images in each category and totally 600 images for each domain.

\begin{figure*}[t]
  \centering
    \subfigure[$\lambda$]{\includegraphics[width=0.32\textwidth]{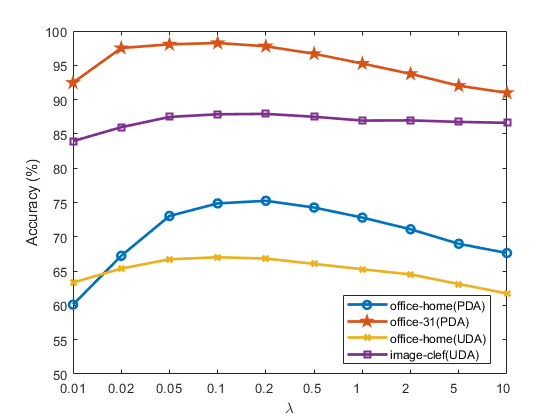}} 
    \subfigure[$\gamma$]{\includegraphics[width=0.32\textwidth]{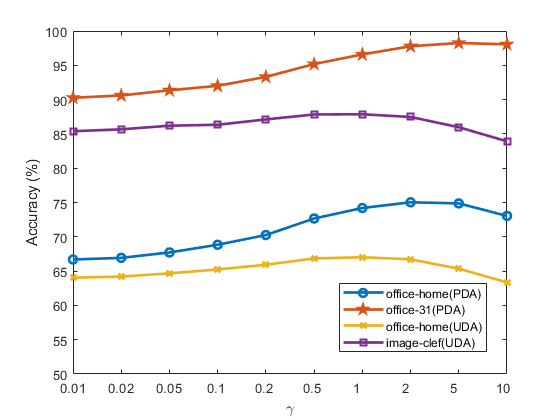}} 
	\subfigure[$\#$neighbor $p$]{\includegraphics[width=0.32\textwidth]{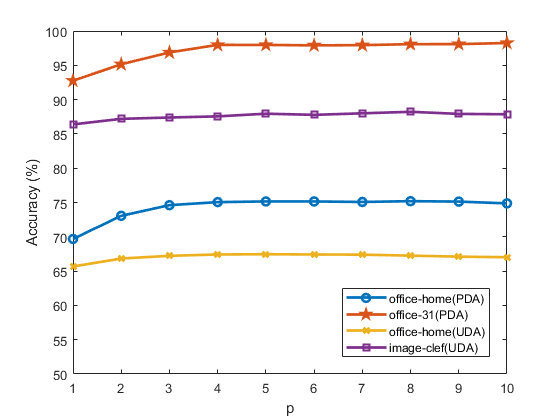}} \\
\caption{classification accuracy w.r.t. $p$, $\lambda$ and $\gamma$, respectively}
	\label{fig2}
	\vspace{0.2in}
\end{figure*}

\begin{figure*}[t]
  \centering
  \includegraphics[width=1\textwidth]{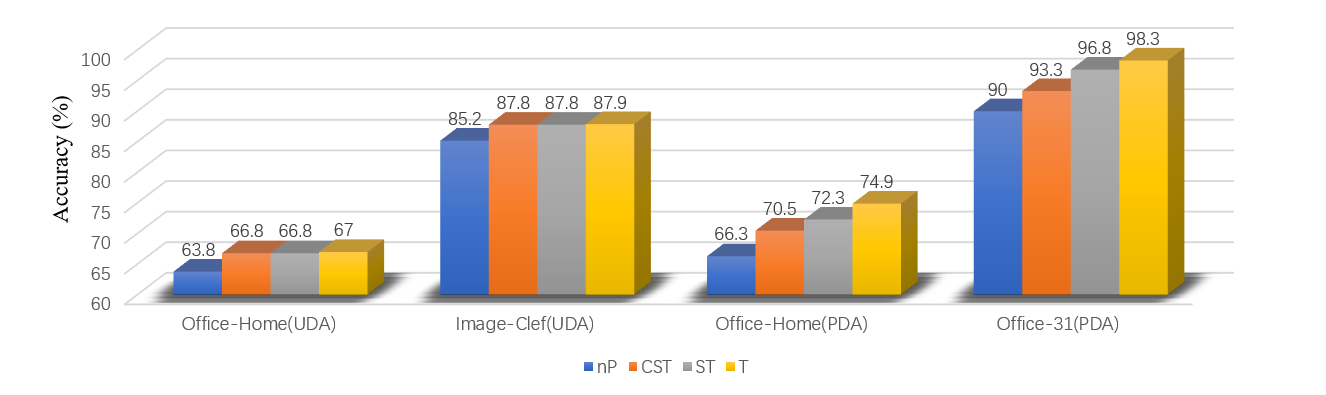}
    \caption{Performance on UDA/PDA with different structure preserving: no structure preserving (nP), inter and intra structure of source and target domain (CST), intra structure of both source and target domain (ST) and intra structure of target domain (T)}
  \label{figab}
\end{figure*}

\subsubsection{Experimental Setup}
We compare IDSP respectively with several UDA \textcolor{black}{baselines}: 1 Nearest Neighbor (1NN), Transfer Component Analysis (TCA) \cite{pan2010domain}, Transfer Joint Matching (TJM) \cite{long2013adaptation}, Correlation Alignment (CORAL) \cite{sun2016return}, Geodesic Flow Kernel (GKF) \cite{gong2012geodesic}, Subspace Alignment (SA) \cite{fernando2013unsupervised}, ResNet50 \cite{he2016deep}, DAN \cite{long2015learning},DANN \cite{ganin2015unsupervised}, Joint Adaptation Networks (JAN) \cite{long2017deep}, Conditional Adversarial Networks (CDAN) \cite{long2018conditional}, MEDA \cite{wang2018visual} and PAS\cite{li2020unsupervised}. The results of the deep-learning-based approaches (e.g., DAN, DANN, JAN and CDAN) are obtained directly from the existing works \cite{ganin2015unsupervised,long2018conditional,long2017deep,long2015learning}. For fair comparison, we use the 2048-dimensional deep feature (extracted using ResNet50 pre-trained on ImageNet) for both IDSP and other shallow UDA approaches. The optimal parameters of all compared methods are set according to their original papers. As for IDSP, we empirically set the hyper-parameters $\lambda=0.1$, $\gamma=1$ and $p=10$ for the UDA setting.

 \subsubsection{Experimental Results}
The classification results of the 12 UDA tasks on \emph{Office-Home} dataset and 6 UDA tasks on \emph{Image-Clef} dataset are given in Table \ref{tab4} and Table \ref{tab5}, respectively. On both two datasets, our approach achieves comparable results with average accuracy of 67.0 \% on \emph{Office-Home} dataset and 87.9\% on \emph{Image-Clef} dataset. Specifically, the IDSP achieves the best/second-best results on 11 tasks at \emph{Office-Home} dataset and achieves the best/second-best on 4 tasks at \emph{Office-31} dataset. The results illustrate the effectiveness and the flexibility of model smoothness for solving UDA. Notably, MEDA incorporates both manifold and domain alignment regularization terms. The superior results of MEDA further show that model smoothness and domain alignment can complement each other since the ideal alignment tends much easier to obtain in UDA.
\begin{table}[]
    \centering
    \caption{Accuracy(\%) of IDSP and IDSP-JDA }
\begin{tabular}{ccc}
\hline\hline
DataSets   & IDSP& IDSP-JDA\\ \hline\hline
Office-Home(UDA) & 67.0  & \textbf{68.0}  \\ 
Image-Clef(UDA) & 87.9   & \textbf{89.3}      \\ 
Office-Home(PDA)  & \textbf{74.9}   & 63.3  \\ Office-31(PDA) & \textbf{98.3}  & 89.3 \\ \hline\hline
\end{tabular}
    
    \label{tab:my_label}
\end{table}
\subsection{Joint Work with Domain Alignment}
 To validate the influence of domain alignment for IDSP, we further conduct experiments by adding an additional joint distribution adaptation (JDA) term \cite{long2014transfer} on UDA and PDA settings. The results are given in Table \ref{tab:my_label}. As we observe, in the UDA setting, the results of (IDSP+JDA) achieve superior results with the accuracy of 68.0 \% on \emph{Office-Home} dataset and 89.3\% on \emph{Image-Clef} dataset.  The results illustrate that domain alignment and IDSP can be complementary with each other at a safe transfer of knowledge (e.g., UDA), \textcolor{black}{which provides 1.0\% and 1.4\% performance gain in \emph{Office-Home} dataset and \emph{Office-31} dataset, respectively, after incorporating domain alignment}. In contrast, in the PDA setting, we can find that \textcolor{black}{ the adaptation performance is significantly decreased after adding domain alignment (i.e., -11.6 \% in \emph{Office-Home} dataset and -9.0 \% in \emph{Office-31} dataset)}. Overall, IDSP in PDA boosts higher extent in performance on \emph{Office-Home} and \emph{Office-31} than IDSP-JDA in UDA on \emph{Office-Home} and \emph{Image-Clef}. More details in the experiment analysis can be found in the Appendix C. The results reveal that it is beneficial to give up domain alignment in PDA at least in more risky settings where perfect alignment is currently hard to be achieved. 


\subsection{Sensitivity Analysis}
The proposed IDSP method involves three hyper-parameters (i.e., $\lambda$ for $l_2$-regularization, $\gamma$ for laplacian regularization and neighbor $p$). To investigate the sensitivity of these hyper-parameters on performance, we conduct experiments on \emph{office-Home}, \emph{Image-Clef} and \emph{Office-31} datasets. Specifically, we run IDSP by searching $\lambda \in\{ 0.01, 0.02, 0.05, 0.1, 0.2, 0.5, 1, 2, 5, 10\}$, $\gamma \in\{ 0.01, 0.02, 0.05, 0.1, 0.2, 0.5, 1, 2, 5, 10\}$ and  $p \in\{ 1,2,3,4,5,6,7,8,9,10\}$. As we observe in Fig. \ref{fig2} (a - c), the IDSP performs robustly and insensitively on both the closed-set UDA and PDA tasks on a wide range of parameter values of $p$, $\lambda$ and $\gamma$. 
 
\subsection{Effectiveness of Intra Domain Structure Preserving}
We verify effectiveness of IDSP by inspecting the impacts of different structure preserving constraints. Specifically, we use no structure preserving (nP, i.e., $\gamma = 0$), conventional manifold structure preserving (CST), intra structure of both source and target domain (ST) and intra structure preserving of target domain (T, i.e., IDSP), whose results on the \emph{Office-Home}, \emph{Image-Clef} and \emph{Office-31} are given in Fig. \ref{figab}. More details are given in the Appendix E. It can be observed that the IDSP outperforms these baselines. From Fig. \ref{figab}, we easily find that the structure preserving can effectively enhance the adaptation ability of the classifier learned from the source domain, which confirms the superiority of model smoothness. Also, we can find that with more source structure information considered (i.e., ST and CST), the performance is reduced, especially in PDA. The reason is that the manifold structure across different domains may vary sharply (especially when the irrelevant categories exist) along with the domain shift, resulting in a negative transfer. The results illustrate the necessity to consider the intra domain structure preserving.

\section{Conclusion\label{s5}}
In this paper, considering the difficulty of the perfect alignment between domains while solving PDA, we endeavor to address the PDA by giving up domain alignment. To achieve it, a novel generalization error bound is first derived and then a theoretically-motivated PDA approach is proposed by enforcing intra domain structure preserving (IDSP). The experimental results demonstrate the effectiveness of the proposed IDSP, which confirms that IDSP scheme can be applicable to enhance the adaptation ability in solving PDA. It should be noted that IDSP naturally benefits the source-free UDA, since it only considers the target structure, which will be further studied in our future work. \textcolor{black}{Besides, since the IDSP can generate reliable labels for most examples, the pseudo label obtained by IDSP can thus serve as a good cold start for the existing pseudo-labeled-based DA methods, when we encounter a new domain adaptation scenario.} In addition, this study also indicates that IDSP and conventional domain alignment can be complementary with each other in the UDA setting. In the end, we would like to emphasize that undoubtedly domain alignment remains important for domain adaptation. Thus, how to obtain harmless or perfect alignment with the guidance of model smoothness is still our next crucial pursuit.
\section*{Acknowledgement}
The authors would like to thank Dr. Jingjing Gu, Dr. Chuanxing Geng and Dr. Yunyun Wang for the proofreading of this manuscript. This work is supported in part by the NSFC under Grant No. 62076124.






%
%
%

\bibliographystyle{ieeetr}
\bibliography{mybibfile}

%
%
%
%
\newpage
\begin{IEEEbiography}[{\includegraphics[width=1in,height=1.25in,clip,keepaspectratio]{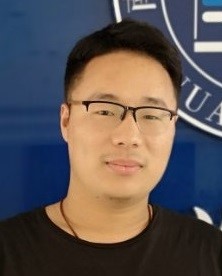}}]{Weikai Li}
received his B.S. degree in Information and Computing Science from Chongqing Jiaotong University in 2015. In 2018, he completed his M.S. degree in computer science and technique at Chongqing Jiaotong University. He is currently pursuing the Ph.D. degree with the College of Computer Science \& Technology, Nanjing University of Aeronautics and Astronautics. His research interests include pattern recognition and machine learning.
\end{IEEEbiography}
\begin{IEEEbiography}[{\includegraphics[width=1in,height=1.25in,clip,keepaspectratio]{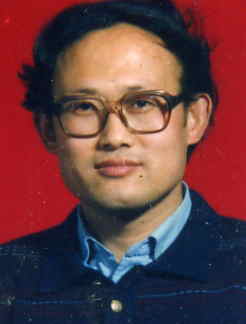}}]{Songcan Chen} 
received his B.S. degree in mathematics from Hangzhou University (now merged into Zhejiang University) in 1983. In 1985, he completed his M.S. degree in computer applications at Shanghai Jiaotong University and then worked at NUAA in January 1986. There he received a Ph.D. degree in communication and information systems in 1997. Since 1998, as a full-time professor, he has been with the College of Computer Science \& Technology at NUAA. His research interests include pattern recognition, machine learning and neural computing. He is also an IAPR Fellow.
\end{IEEEbiography}

%
%
%




\end{document}


\appendices

\setcounter{theorem}{0}   
\setcounter{corollary}{0}   



\section{Generalization Error}
\begin{theorem}
\label{the2}
 Given two distributions $\mathbb{P}$ and $\mathbb{Q}$, if a model $f$ is $2r$-cover with $\epsilon$ smoothness over distributions $\mathbb{P}$ and $\mathbb{Q}$, with probability at least $ 1 - \theta$, we have:
\end{theorem}
\begin{equation}
\label{eq13}
\begin{aligned}
   \mathcal{E}_{\mathbb{Q}}\left(f\right) 
   \leq  \mathcal{E}_{\mathbb{P}}\left(f\right) +2&\epsilon+ 2M\mathrm{TV}(\mathbb{P},\mathbb{Q})\\
    +  M&\sqrt{\frac{\left(2 d\right)^{\frac{2 \epsilon^{2} D}{r^{2}}+1} \log 2+2 \log \left(\frac{1}{\theta}\right)}{m}}\\
      +  M&\sqrt{\frac{\left(2 d\right)^{\frac{2 \epsilon^{2} D}{r^{2}}+1} \log 2+2 \log \left(\frac{1}{\theta}\right)}{n}}\\
       +  M&\sqrt{\frac{\log(1/\theta)}{2m}}.
\end{aligned}
\end{equation}

\noindent
\textbf{Proof}
\ref{eq13},eq14
\noindent
Let $B_{\mathrm{W}}\left(\mathbb{P}, r\right)=\left\{P: \mathrm{W}_\infty\left(\mathbb{P}, P\right) \leq r\right\}$, where $\mathrm{W}_\infty$ is the $\infty$-th Wasserstein-distance \cite{villani2009optimal}. Let $\mathbb{P}_r = \arg\min_{p\in B_{\mathrm{W}}\left(\mathbb{P}, r\right)}\mathcal{E}_{p}\left(f\right) $ and $\mathbb{Q}_r = \arg\min_{q\in B_{\mathrm{W}}\left(\mathbb{Q}, r\right)}\mathcal{E}_{q}\left(f\right) $
\begin{equation}
\label{eq14}
\begin{aligned}
   \mathcal{E}_{\mathbb{Q}}\left(f\right) = &\mathcal{E}_{\mathbb{Q}}\left(f\right) - \mathcal{E}_{\mathbb{Q}_r}\left(f\right) +\mathcal{E}_{\mathbb{Q}_r}\left(f\right)- \mathcal{E}_{\mathbb{P}}\left(f\right) +\mathcal{E}_{\mathbb{P}}\left(f\right) \\
   \leq & \mathcal{E}_{\mathbb{P}}\left(f\right) +|\mathcal{E}_{\mathbb{Q}_r}\left(f\right) - \mathcal{E}_{\mathbb{Q}}\left(f\right) |\\
   &+ |\mathcal{E}_{\mathbb{Q}_r}\left(f\right)-\mathcal{E}_{\mathbb{P}}\left(f\right)| \\
   \leq & \mathcal{E}_{\mathbb{P}}\left(f\right) + |\mathcal{E}_{\mathbb{Q}_r}\left(f\right) - \mathcal{E}_{\mathbb{Q}}\left(f\right) | \\
   &+ |\mathcal{E}_{\mathbb{Q}_r}\left(f\right)-\mathcal{E}_{\mathbb{P}_r}\left(f\right)|+|\mathcal{E}_{\mathbb{P}_r}\left(f\right)-\mathcal{E}_{\mathbb{P}}\left(f\right)|. 
\end{aligned}
\end{equation} 
Then, according to the \textbf{Theorems} 1 and 4 in a recent study \cite{yi2021improved}, we have

\begin{equation}
\label{eq15}
\begin{aligned}
    |\mathcal{E}_{\mathbb{P}_r}\left(f\right) - \mathcal{E}_{\mathbb{P}}\left(f\right) | \leq& \epsilon\\
    +  M&\sqrt{\frac{\left(2 d\right)^{\frac{2 \epsilon^{2} D}{r^{2}}+1} \log 2+2 \log \left(\frac{1}{\theta}\right)}{m}},
\end{aligned}
\end{equation}

\begin{equation}
\label{eq16}
\begin{aligned}
    |\mathcal{E}_{\mathbb{Q}_r}\left(f\right) - \mathcal{E}_{\mathbb{Q}}\left(f\right) | \leq& \epsilon\\
    +  M&\sqrt{\frac{\left(2 d\right)^{\frac{2 \epsilon^{2} D}{r^{2}}+1} \log 2+2 \log \left(\frac{1}{\theta}\right)}{n}},
\end{aligned}
\end{equation}
and
\begin{equation}
\label{eq17}
    |\mathcal{E}_{\mathbb{Q}_r}\left(f\right)-\mathcal{E}_{\mathbb{P}_r}\left(f\right)| \leq 2M\mathrm{TV}(\mathbb{P},\mathbb{Q})    +  M\sqrt{\frac{\log(1/\theta)}{2m}}.
\end{equation}
By plugging the Eqs. \ref{eq15}, \ref{eq16} and \ref{eq17} into Eq. \ref{eq14}, we have :

\begin{equation}
\label{eq19}
\begin{aligned}
   \mathcal{E}_{\mathbb{Q}}\left(f\right) 
   \leq  \mathcal{E}_{\mathbb{P}}\left(f\right) +2&\epsilon+ 2M\mathrm{TV}(\mathbb{P},\mathbb{Q})\\
    +  M&\sqrt{\frac{\left(2 d\right)^{\frac{2 \epsilon^{2} D}{r^{2}}+1} \log 2+2 \log \left(\frac{1}{\theta}\right)}{m}}\\
      +  M&\sqrt{\frac{\left(2 d\right)^{\frac{2 \epsilon^{2} D}{r^{2}}+1} \log 2+2 \log \left(\frac{1}{\theta}\right)}{n}}\\
       +  M&\sqrt{\frac{\log(1/\theta)}{2m}}.
\end{aligned}
\end{equation} 

\noindent
\textbf{Q.E.D}

\subsection{Why TV-divergence}
\textcolor{black}{Current research has revealed the incoherence between the widely-adopted empirical domain adversarial training and its generally-assumed theoretical counterpart based on $\mathcal{H}$-divergence \cite{shui2020beyond}. While in fact, the domain adversarial training methods align the target domain to the source data by cheating the domain classifier, which is proven to be an approximation of KL-divergence \cite{pan2020loss}. Note that TV distance can be upper bounded in terms of KL-divergence as $TV(\mathbb{P},\mathbb{Q})\leq\sqrt{KL(\mathbb{P},\mathbb{Q})/{2\log e}}$ \cite{nielsen2018guaranteed}. Thus, we conduct the TV distance to analyze the new target generalization error bound.}
\newpage
\section{Representer Theorem}
\begin{theorem}
\label{the1}
 Representer Theorem: The parameter $\mathbf{W}^{*}=\left[\mathbf{w}_{1}^{*}, \cdots, \mathbf{w}_{h}^{*}\right]$ for the optimized solution $f$ can be expressed in terms of the cross-domain labeled and unlabeled examples, 
\end{theorem}
\begin{equation}
\label{eq7}
    f(\mathbf{x})=\sum_{i=1}^{n+m} \alpha_{i} K\left(\mathbf{x}_{i}, \mathbf{x}\right) \quad \text { and } \quad \mathbf{w}=\sum_{i=1}^{n+m} \alpha_{i} \phi\left(\mathbf{x}_{i}\right),
\end{equation}
where $K$ is a kernel induced by $\phi$, $\alpha_i$ is a coefficient.

\noindent
\textbf{Proof} It is easy to prove by contradiction.
Denote span ($\phi(\mathbf{x}_i), 1\leq i\leq n+m$) as the linear space spanned by vectors $\phi(\mathbf{x}_i): 1\leq i\leq n+m$. Then, each $\mathbf{w}_j$ can be expressed as
\begin{equation}
    \mathbf{w}_{j}=\mathbf{w}_{j}^{\|}+\mathbf{w}_{j}^{\perp},
\end{equation}
where $\mathbf{w}_{j}^{\|}$ is the component along the linear space span($\phi(\mathbf{x}_i), 1\leq i\leq n+m$) and $\mathbf{w}_{j}^{\perp}$ is the component along its orthogonal complement space. Let the optimal solution
\begin{equation}
    \mathbf{W}^{*}=\mathbf{W}^{* \|}+\mathbf{W}^{* \perp},
\end{equation}
where $\mathbf{W}^{* \|}=\left[\mathbf{w}_{1}^{* \|}, \mathbf{w}_{2}^{* \|}, \cdots, \mathbf{w}_{m}^{* \|}\right]$ and $\mathbf{W}^{* \perp}=\left[\mathbf{w}_{1}^{* \perp}, \mathbf{w}_{2}^{* \perp}, \cdots, \mathbf{w}_{m}^{* \perp}\right]$.  Then, we easily have $\left(\mathbf{W}^{* \perp}\right)^{T} \mathbf{W}^{* \|}=\mathbf{0}$.  We assume that $\mathbf{W}^{* \perp}\neq 0$, and denote $\mathcal{J(\cdot)}$ as the objective value of IDSP, then we have :
\begin{equation}
\begin{aligned}
     \mathcal{J}(\mathbf{W}^*)=&\sum_{i=1}^{n}\left(y_{i}-f\left(\mathbf{x}_{i}\right)\right)^{2}+\lambda\|f\|_{K}^{2}\\
     &+\gamma \sum_{i, j=1}^{n+m} f\left(\mathbf{x}_{i}\right) L_{i j}f\left(\mathbf{x}_{j}\right)\\
     =&\sum_{i=1}^{n}\left(y_{i}-\mathbf{w} ^{*\intercal}  \phi(\mathbf{x}_i)\right)^{2}+\lambda\operatorname{tr}\left(\mathbf{w} ^{*\intercal} \mathbf{w} ^{*}\right)\\
     &+\gamma \sum_{i, j=1}^{n+m} \mathbf{w} ^{*\intercal} \phi(\mathbf{x}_i) L_{i j}\mathbf{w} ^{*\intercal} \phi(\mathbf{x}_j)\\
     =&\sum_{i=1}^{n}\left(y_{i}-\mathbf{w} ^{*\|\intercal}  \phi(\mathbf{x}_i)\right)^{2}\\
     &+\lambda\left(\operatorname{tr}\left(\mathbf{w} ^{*\|\intercal} \mathbf{w} ^{*\|}\right)+\operatorname{tr}\left(\mathbf{w} ^{*\perp\intercal} \mathbf{w} ^{*\perp}\right)\right)\\
     &+\gamma \sum_{i, j=1}^{n+m} \mathbf{w} ^{*\|\intercal} \phi(\mathbf{x}_i) L_{i j}\mathbf{w} ^{*\|\intercal} \phi(\mathbf{x}_j)\\
     &+\gamma \sum_{i, j=1}^{n+m} \mathbf{w} ^{*\perp\intercal} \phi(\mathbf{x}_i) L_{i j}\mathbf{w} ^{*\perp\intercal} \phi(\mathbf{x}_j).
\end{aligned}
\end{equation}
While  $\mathbf{W}^{* \perp}\neq 0$, and $\mathbf{L}$ is the Laplacian matrix, we easily have $\mathcal{J(\mathbf{w} ^{*\|})}\leq \mathcal{J(\mathbf{w} ^{*})}$ .Thus, $\mathbf{W}^{* \|}$ is more optimal than $\mathbf{W}^{*}$, which is contradictory with the fact. This yields the  $\mathbf{W}^{*} = \mathbf{W}^{* \|}$.

\noindent
\textbf{Q.E.D}

\newpage

\section{Joint Work with Domain Alignment}
Notably, the risk of domain alignment tends to be lower in UDA. We further conduct experiments by adding an additional joint distribution adaptation (JDA) term \cite{long2014transfer}. The entire objective is given as follows:
\begin{equation}\label{eq22}
\begin{aligned}
 f=\underset{f \in \mathcal{H}_{K}}{\arg \min }& \sum_{i=1}^{n}\left(y_{i}-f\left(\mathbf{x}_{i}\right)\right)^{2}+\lambda\|f\|_{K}^{2}\\
 +&\gamma \sum_{i, j=1}^{n+m} f\left(\mathbf{x}_{i}\right) L_{i j}f\left(\mathbf{x}_{j}\right) +\eta\mathcal{R}_{JDA}(f),
\end{aligned}
\end{equation}
where $\mathcal{R}_{JDA}(f)$ is the JDA term, which is denoted as :
\begin{equation}
\begin{aligned}
     \mathcal{R}_{JDA}(f) &= \left\|\frac{1}{n} \sum_{i=1}^{n} f\left(\mathbf{x}_{i}\right)-\frac{1}{m} \sum_{j=n+1}^{n+m} f\left(\mathbf{x}_{j}\right)\right\|_{\mathcal{H}}^{2}\\
     +&\left\|\frac{1}{n^{(c)}} \sum_{\mathbf{x}_{i} \in \mathcal{D}_{s}^{(c)}} f\left(\mathbf{x}_{i}\right)-\frac{1}{m^{(c)}} \sum_{\mathbf{x}_{j} \in \mathcal{D}_{t}^{(c)}} f\left(\mathbf{x}_{j}\right)\right\|_{\mathcal{H}}^{2},
\end{aligned}
\end{equation}
where $\mathcal{D}_{s}^{(c)}=\left\{\mathbf{x}_{i}: \mathbf{x}_{i} \in \mathcal{D}_{s} \wedge y\left(\mathbf{x}_{i}\right)=c\right\}$ is the set of the source data belonging to class $c$.and $n^{(c)}=\left|\mathcal{D}_{S}^{(c)}\right|$. Correspondingly, $\mathcal{D}_{t}^{(c)}= \left\{\mathbf{x}_{j}: \mathbf{x}_{j} \in \mathcal{D}_{t} \wedge \hat{y}\left(\mathbf{x}_{j}\right)=c\right\}$, where $\hat{y}(\mathbf{x}_j)$is the pseudo (predicted) label of $\mathbf{x}_j$ and $m^{(c)}=\left|\mathcal{D}_{t}^{(c)}\right|$.
\subsection*{Learning Algorithm}
\begin{theorem}
\label{the3}
The minimizer of optimization problem \ref{eq22} admits an expansion 
\end{theorem}
\begin{equation}
\label{eqt}
    f(\mathbf{x})=\sum_{i=1}^{n+m} \alpha_{i} K\left(\mathbf{x}_{i}, \mathbf{x}\right) \quad \text { and } \quad \mathbf{w}=\sum_{i=1}^{n+m} \alpha_{i} \phi\left(\mathbf{x}_{i}\right),
\end{equation}
where $K$ is a kernel induced by $\phi$, $\alpha_i$ is a coefficient.

\noindent
\textbf{Proof} The proof is similar to Theorem \ref{the1}. Hence, we omit the proof for Theorem \ref{the3}.

By incorporating Eq.\ref{eqt} into Eq. \ref{eq22}, we obtain the following objective:
\begin{equation}
    \begin{aligned}
\alpha=& \underset{\alpha \in \mathbb{R}^{n+m}}{\arg \min }\left\|\left(\mathbf{Y}-\boldsymbol{\alpha}^{\mathrm{T}} \mathbf{K}\right) \mathbf{V}\right\|_{F}^{2} \\
&+\operatorname{tr}\left(\lambda \boldsymbol{\alpha}^{\mathrm{T}} \mathbf{K} \boldsymbol{\alpha}+\boldsymbol{\alpha}^{\mathrm{T}} \mathbf{K}(\gamma\mathbf{L}+\eta\mathbf{M}) \mathbf{K} \boldsymbol{\alpha}\right),
\end{aligned}
\end{equation}
where $\mathbf{V}$ is the label indicator matrix with $\mathbf{V}_{ii}=1$ if $i \in \mathcal{D}_s$, otherwise$\mathbf{V}_{ii}=0$.  $\mathbf{M}=\sum_{c=0}^CM_c$ and $M_c,c\in\{0,1,\cdots,C\}$ are the MMD matrices which can be computed as:
\begin{equation}\label{eqm}
\left(M_{c}\right)_{i j}=\left\{\begin{array}{ll}
\frac{1}{n^{(c)} n^{(c)}}, & \mathbf{x}_{i}, \mathbf{x}_{j} \in \mathcal{D}_{s}^{(c)} \\
\frac{1}{m^{(c)} m^{(c)}}, & \mathbf{x}_{i}, \mathbf{x}_{j} \in \mathcal{D}_{t}^{(c)} \\
\frac{-1}{n^{(c)} m^{(c)}}, & \left\{\begin{array}{l}
\mathbf{x}_{i} \in \mathcal{D}_{s}^{(c)}, \mathbf{x}_{j} \in \mathcal{D}_{t}^{(c)} \\
\mathbf{x}_{j} \in \mathcal{D}_{s}^{(c)}, \mathbf{x}_{i} \in \mathcal{D}_{t}^{(c)}
\end{array}\right. \\
0, & \text { otherwise }
\end{array}\right.
\end{equation}
For clarity, we can also compute $\mathbf{M}_0$ with Eq. \ref{eqm} if substituting $n^{(0)}=n, m^{(0)}=m, \mathcal{D}_{s}^{(0)}=\mathcal{D}_{s}, \mathcal{D}_{t}^{(0)}=\mathcal{D}_{t}$
Setting derivative of objective function as 0 leads to:

\begin{equation}
\label{eq12}
    \boldsymbol{\alpha}=((\mathbf{V}+\gamma \mathbf{L}+\eta\mathbf{M}) \mathbf{K}+\lambda \mathbf{I})^{-1} \mathbf{V} \mathbf{Y}^{\mathbf{T}} .
\end{equation}
The learning algorithms are summarized in Algorithm \ref{alg1}.

\begin{algorithm}[t]
    \caption{Learning algorithm for JDA-IDSP}
    \label{alg1} 
    \begin{algorithmic}[1]
    \REQUIRE ~~\\
    $n$ source labeled datasets $\mathcal{D}_s=\{\mathbf{x}^s_i,y_i^s\}_{i=1}^n$\\
    $m$ target unlabeled datasets $\mathcal{D}_t=\{\mathbf{x}^t_i\}_{i=n+1}^{m+n}$\\
    Hyper-parameters $\lambda$, $\gamma$,$p$ and $\eta$;\\ 

    \ENSURE ~~\\ 
    Predictive Classifier $f$
    \STATE Initialize the pseudo label $\hat{y}$
    \STATE Calculate the graph Laplacian $\mathbf{L}$ 
    \STATE Construct kernel $\mathbf{K}$  by a specific kernel function
    \WHILE {not converge}
    \STATE Construct MMD matrix $\mathbf{M}$;
    \STATE Compute $\mathbf{\alpha}$ ;
    \STATE Update the pseudo label $\hat{y}$
    \ENDWHILE 
    \STATE \textbf{Return} Classifier $f$ .
    \end{algorithmic}
\end{algorithm}

\begin{figure}[t]
  \centering
    \subfigure[\textbf{Office-Home}(UDA)]{\includegraphics[width=0.23\textwidth]{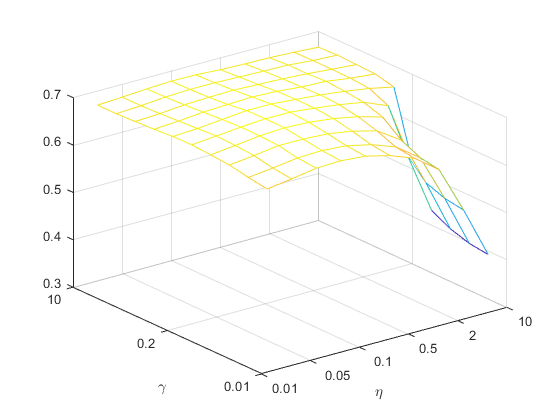}} 
    \subfigure[\textbf{Image-Clef}(UDA)]{\includegraphics[width=0.23\textwidth]{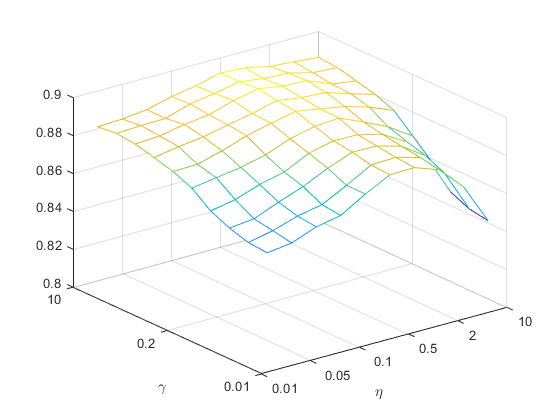}} \\
	\subfigure[\textbf{Office-Home}(PDA)]{\includegraphics[width=0.23\textwidth]{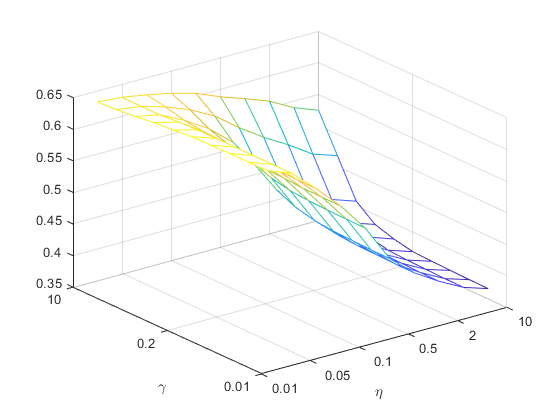}}
	\subfigure[\textbf{Office-31}(PDA)]{\includegraphics[width=0.23\textwidth]{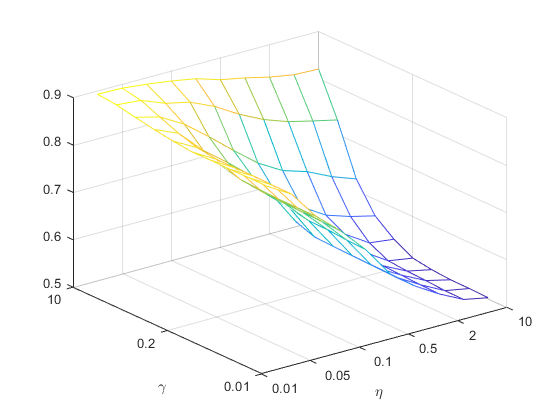}}\\
\caption{classification accuracy w.r.t.$\eta$ and $\gamma$, respectively}
	\label{fig2}
	\vspace{0.2in}
\end{figure}

\begin{figure*}[t]{
  \centering
    \subfigure[\textbf{Office-Home}(UDA)]{\includegraphics[width=0.8\textwidth]{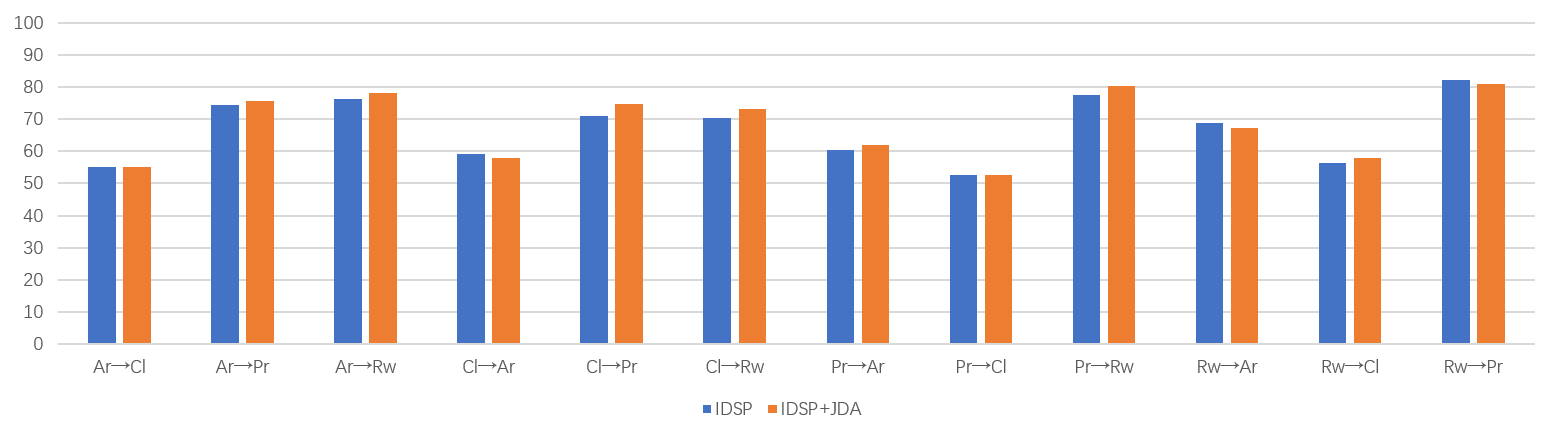}} \\
    \subfigure[\textbf{Image-Clef}(UDA)]{\includegraphics[width=0.4\textwidth]{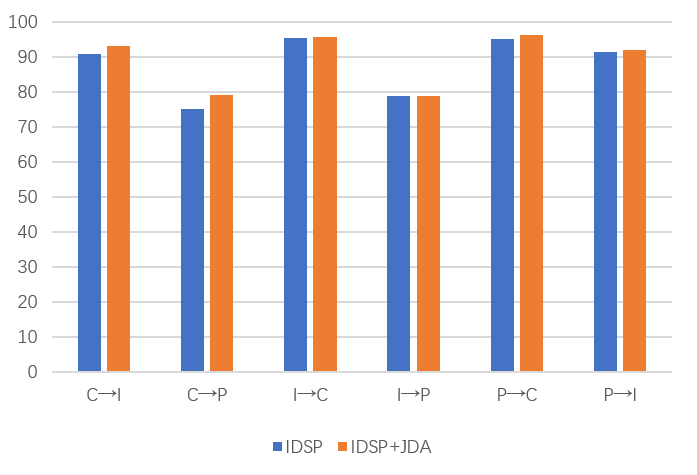}} 
	\subfigure[\textbf{Office-31}(PDA)]{\includegraphics[width=0.4\textwidth]{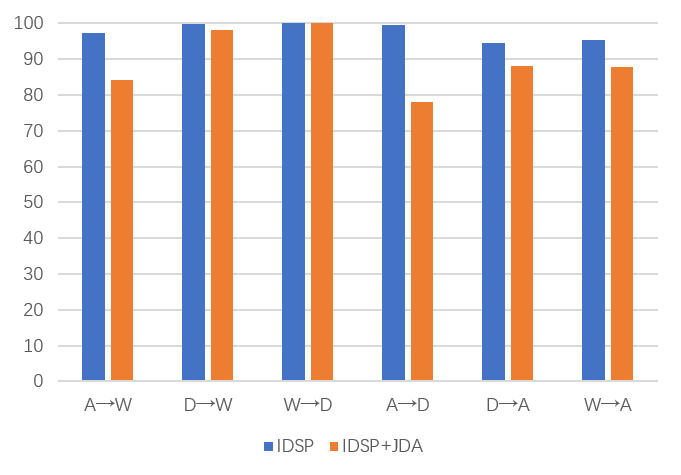}}\\
	\subfigure[\textbf{Office-Home}(PDA)]{\includegraphics[width=0.8\textwidth]{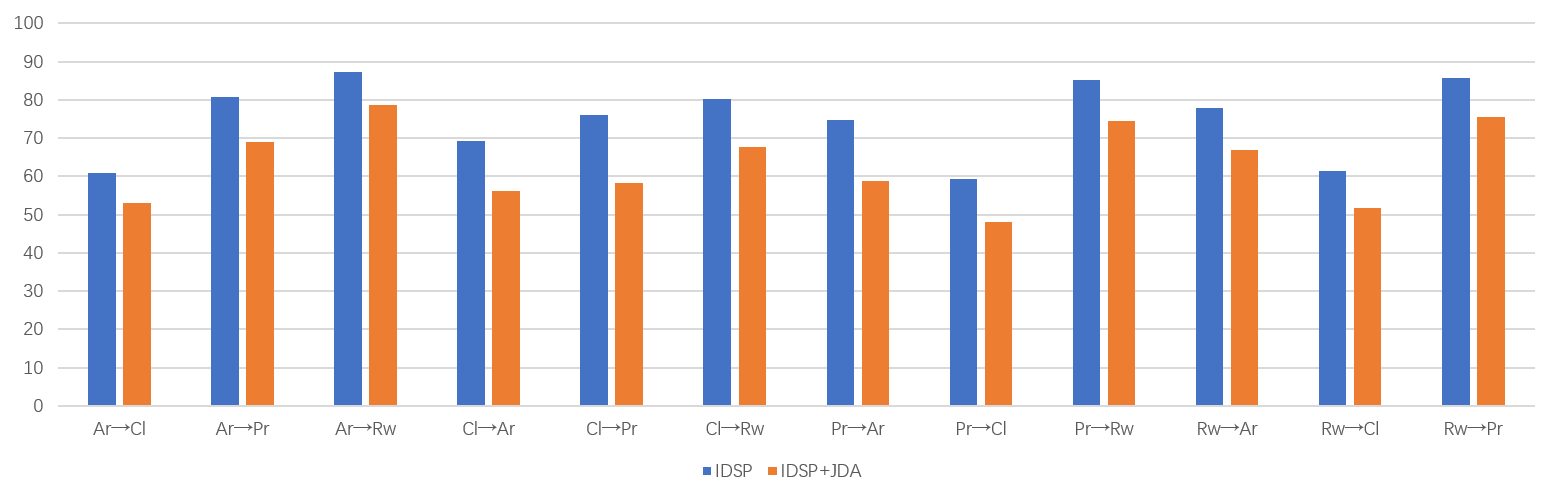}}
\caption{classification accuracy of IDSP and IDSP+JDA}
	\label{fig3}
	\vspace{0.2in}}
\end{figure*}

\subsection*{Sensitivity Analysis}
The proposed IDSP-JDA method involves one additional more hyper-parameters (i.e., $\eta$). To investigate the sensitivity of the smoothness and the domain alignment, we conduct experiments on \emph{office-Home}, \emph{Image-Clef} and \emph{Office-31} datasets. Specifically, we run IDSP by searching $\gamma \in\{ 0.01, 0.02, 0.05, 0.1, 0.2, 0.5, 1, 2, 5, 10\}$, $\eta \in\{ 0.01, 0.02, 0.05, 0.1, 0.2, 0.5, 1, 2, 5, 10\}$ with $\lambda =0.1$ and $p=10$. As we observe in Fig. \ref{fig2}, the IDSP performs robustly and insensitively on both the closed-set UDA on a wide range of parameter values of $\eta$ and $\gamma$. To sum up, the performance of IDSP stays robust with a wide range of regularization parameter choice. They can be selected without knowledge in real UDA applications. In contrast, the performance will be decrease with $\eta$ increased in PDA, which illustrates that the mis-alignment does more harm than good in the PDA setting.
\begin{table}[]
    \centering
    \caption{Accuracy(\%) of IDSP and IDSP-JDA }
\begin{tabular}{ccc}
\hline\hline
DataSets   & IDSP& IDSP-JDA\\ \hline\hline
Office-Home(UDA) & 67.0  & 68.0  \\ 
Image-Clef(UDA) & 87.9   & 89.3      \\ 
Office-Home(PDA)  & 74.9   & 63.3  \\ Office-31(PDA) & 98.3  & 89.3 \\ \hline\hline
\end{tabular}
    
    \label{tab:my_label}
\end{table}

\subsection*{Classification Results}
The classification results of IDSP and IDSP+JDA (($\gamma=2,\eta=0.5$ for UDA and $\gamma=10,\eta=0.01$ for PDA) are given in Fig. \ref{fig3} and Table \ref{tab:my_label}. As we can observe in Fig. \ref{fig3} and Table \ref{tab:my_label}, with JDA incorporates in learning classier, the performance are increased in almost all tasks in the UDA setting \textcolor{black}{(i.e., 1.6\% in \emph{Office-Home} dataset and 1.4\% in \emph{Office-31} dataset)}. In contrast, the performance are significantly decreased in all tasks in the PDA setting, \textcolor{black}{the adaptation performance are significantly decreased after adding domain alignment (i.e., -11.6\% in \emph{Office-Home} dataset and -9.0\% in \emph{Office-31} dataset)}. The results illustrate that the domain alignment may introduce negative transfer in PDA, since PDA do not satisfy the assumption of domain alignment. In contrast, model smoothness can do well on both PDA and UDA settings.
\newpage
\section{The Extension of Model Smoothness}
\textcolor{black}{The IDSP is only a naive empirical validation for model smoothness by assuming the data lies in the Riemann manifold space, which is incorporated into a kernel regression model. In fact, there are several extensions of model smoothness in addressing PDA. In this part, we just list a few extensions.}
\subsection{IDSP on Deep model}
\textcolor{black}{Specifically, the proposed IDSP can be easily adapted to end-to-end format by incorporating the IDSP loss to the existing objective function, which can be solved by error back-propagation \cite{tomar2014manifold,weston2012deep,kejani2020graph}, ADMM-BP \cite{li2019deep} and point integral methods \cite{zhu2018ldmnet}. In addition, we can  utilize the memory bank to store the neighborhood information \cite{dwibedi2021little} to realize IDSP in the end-to-end mode.Besides, we can also directly concatenate the deep neural network and kernel machine \cite{ tang2013deep} to incorporates IDSP into the deep model. Further, since the IDSP can generate reliable labels for most examples, the pseudo label obtained by IDSP can thus serve as a good cold start for the existing pseudo-labeled-based DA methods, when we encounter a new domain adaptation scenario.}

\subsection{Data Disturbance for Model Smoothness}
\textcolor{black}{ To realize the model smoothness, we can also incorporate other consistency regularization terms, such as implicit data disturbance (e.g., Jocabian norm ) and explicit data disturbance (e.g., robust constraint) by assuming the data lies in the Euclidean space. More details are listed as follows:}
\subsubsection{Explicit Data Disturbance for model Smoothness}

\textcolor{black}{The most direct way for optimizing the model smoothness/consistence is to enforce classifier consistency within the norm-ball neighborhood of each sample by explicitly disturbing the data. In fact, several attempts are motivated to address DA by heuristically incorporating an additional model smoothness constraint into their objectives. For example,, Shu et al, \cite{shu2018dirt} and Lee et al, \cite{lee2019drop} incorporate the locally-Lipschitz constraint via
virtual adversarial training enforces classifier consistency, which enforces classifier consistency within the norm-ball neighborhood by minimizing the KL divergence between the sample and the worst-case disturbance. The objective function $\mathcal{L}_{\mathcal{J}_1}$ can be formulated to:}
\begin{equation}
\mathcal{L}_{\mathcal{J}_1}=\min_f\max _{\|r\| \leq c} \mathrm{D}_{\mathrm{KL}}\left(f(x) \| f(x+r)\right)
\end{equation}
\textcolor{black}{Moreover, Yang et al, \cite{yang2020phase} minimizes the difference between the original and the transformed phases and increases the phase consistency of the learned model, whose objective function $\mathcal{L}_{\mathcal{J}_2}$ is given as }
\begin{equation}
\mathcal{L}_{\mathcal{J}_2}=min_f \frac{-\left\langle f(x), f(T(x))_{j}\right\rangle}{\left\|f(x)_{j}\right\|_{2} \cdot\left\|f(T(x))_{j}\right\|_{2}}
\end{equation}
\textcolor{black}{where $T(.)$ is a transformation operator.}

\textcolor{black}{In addition, Mishra1 et al, \cite{mishra2021surprisingly} designs a consistency regularization, which promotes the model to produce the same output for both an input image and a perturbed version. The objective function $\mathcal{L}_{\mathcal{J}_3}$  is as follows} 
\begin{equation}
\mathcal{L}_{\mathcal{J}_3}=\min_f CE \left(f(x),f(x+r)\right)
\end{equation}
where $CE\left(p, q\right)=- p \log q$ is the cross-entropy. 
\subsubsection{Implicit Data Disturbance for Model Smoothness}
\textcolor{black}{Moreover, we can also implicitly disturb the data to optimize the model smoothness. Along this line, we relax the model smoothness for simplicity, and obtain the following objective:}

\begin{equation}
\label{eq6}
    \frac{1}{\sigma^{2}} \sum_{i=1}^{n_{t}} \mathbb{E}_{r}\left(f\left(\mathbf{x}_{i}+r\right)-f\left(\mathbf{x}_{i}\right)\right)^{2},
\end{equation}
where $r \sim N\left(0, \sigma^{2} \mathbf{I}\right)$. By first-order Taylor expansion, and let $\mathbf{J}(\mathbf{x})=\frac{\partial \mathbf{f}}{\partial \mathbf{x}} \in R^{K \times D}$, we have
\begin{equation}
    f(\mathbf{x}+r)=f(\mathbf{x})+\mathbf{J}(\mathbf{x}) r+\boldsymbol{o}(r).
\end{equation}
\textcolor{black}{Omitting the high-order terms, Eq. \ref{eq6} can be reformulated to the following JN regularizer ($\mathcal{L}_{\mathcal{J}_4}$):}
\begin{equation}
\label{eq8}
\begin{aligned}
\mathcal{L}_{\mathcal{J}_4}&=\frac{1}{\sigma^{2}} \sum_{i=1}^{n_{t}} \mathbb{E}_{r}\left(f \left(\mathbf{x}_{i}\right.\right.\left.+r)-f\left(\mathbf{x}_{i}\right)\right)^{2} \\
&=\frac{1}{\sigma^{2}} \sum_{i=1}^{n_{t}} \mathbb{E}_{r}\left\|\mathbf{J}\left(\mathbf{x}_{i}\right) r\right\|^{2} \\
&=\sum_{i=1}^{n_{t}} \operatorname{tr}\left[\mathbf{J}\left(\mathbf{x}_{i}\right)^{T} \mathbf{J}\left(\mathbf{x}_{i}\right) \frac{1}{\sigma^{2}} \mathbb{E}_{r}\left[r r^{T}\right]\right] \\
&=\sum_{i=1}^{n_{t}} \operatorname{tr}\left[\mathbf{J}\left(\mathbf{x}_{i}\right)^{T} \mathbf{J}\left(\mathbf{x}_{i}\right) \frac{1}{\sigma^{2}} \sigma^{2} \mathbf{I}\right] \\
&=\sum_{i=1}^{n_{t}}\left\|\mathbf{J}\left(\mathbf{x}_{i}\right)\right\|_{F}^{2}.
\end{aligned}
\end{equation}

\begin{table*}[]
\centering
\caption{Classification accuracies (\%) on Office-Home dataset for closed-set DA (ResNet-50).}
\begin{tabular}{cccccccccccccc}
\hline\hline
Methods   & Ar$\rightarrow$Cl & Ar$\rightarrow$Pr & Ar$\rightarrow$Rw & Cl$\rightarrow$Ar & Cl$\rightarrow$Pr & Cl$\rightarrow$Rw & Pr$\rightarrow$Ar & Pr$\rightarrow$Cl & Pr$\rightarrow$Rw & Rw$\rightarrow$Ar & Rw$\rightarrow$Cl &      Rw$\rightarrow$Pr & AVE\\\hline\hline
ResNet    & 34.9  & 50.0 & 58.0 & 37.4 & 41.9 & 46.2 & 38.5 & 31.2 & 60.4 & 53.9 & 41.2 & 59.9 & 46.1 \\
DANN      & 45.6  & 59.3 & 70.1 & 47.0 & 58.5 & 60.9 & 46.1 & 43.7 & 68.5 & 63.2 & 51.8 & 76.8 & 57.6 \\
JAN       & 45.9  & 61.2 & 68.9 & 50.4 & 59.7 & 61.0 & 45.8 & 43.4 & 70.3 & 63.9 & 52.4 & 76.8 & 58.3 \\
CDAN      & 50.7  & 70.6 & 76.0 & 57.6 & 70.0 & 70.0 & 57.4 & 50.9 & 77.3 & 70.9 & 56.7 & 81.6 & 65.8 \\
ABAS      & 51.5  & 71.7 & 75.5 & 59.8 & 69.5 & 69.5 & 59.8 & 47.1 & 77.7 & 70.6 & 55.2 & 80.2 & 65.7 \\
TADA      & 53.1  & 72.3 & 77.2 & 59.1 & 72.1 & 72.1 & 59.7 & 53.1 & 78.4 & 72.4 & 60.0 & 82.9 & 67.6 \\
CADA-A    & 56.9  & 75.4 & 80.2 & 61.7 & 74.9 & 74.9 & 62.9 & 54.4 & 80.9 & 74.3 & 61.1 & 84.4 & 70.1 \\
DCAN      & 54.5  & 75.7 & 81.2 & 67.4 & 74.0 & 76.3 & 67.4 & 52.7 & 80.6 & 74.1 & \textbf{59.1} & 83.5 & 70.5 \\
SHOT      & 55.9  & 77.8 & 80.7 & 66.9 & 77.3 & 77.1 & 66.4 & 53.6 & 81.5 & 72.1 & 57.2 & 83.0 & 70.8 \\
SHOT+IDSP & \textbf{58.8}  & \textbf{79.7} & \textbf{81.6} & \textbf{69.3} & \textbf{80.0} & \textbf{79.1} & \textbf{68.2} & \textbf{55.1} & \textbf{82.2} & \textbf{76.4} & 59.0 & \textbf{84.6} & \textbf{72.8}\\
\hline\hline
\end{tabular}
    \label{tabs2}
\end{table*}

\begin{table*}[]
\centering
\caption{Classification accuracies (\%) on Office-Home dataset for partial-set DA (ResNet-50).}
\begin{tabular}{cccccccccccccc}
\hline\hline
Methods   & Ar$\rightarrow$Cl & Ar$\rightarrow$Pr & Ar$\rightarrow$Rw & Cl$\rightarrow$Ar & Cl$\rightarrow$Pr & Cl$\rightarrow$Rw & Pr$\rightarrow$Ar & Pr$\rightarrow$Cl & Pr$\rightarrow$Rw & Rw$\rightarrow$Ar & Rw$\rightarrow$Cl &      Rw$\rightarrow$Pr & AVE\\\hline\hline
ResNet    & 46.3  & 67.5 & 75.9 & 59.1 & 59.9 & 62.7 & 58.2 & 41.8 & 74.9 & 67.4 & 48.2 & 74.2 & 61.3 \\
IWAN      & 53.9  & 54.5 & 78.1 & 61.3 & 48.0 & 63.3 & 54.2 & 52.0 & 81.3 & 76.5 & 56.8 & 82.9 & 63.6 \\
SAN       & 44.4  & 68.7 & 74.6 & 67.5 & 65.0 & 77.8 & 59.8 & 44.7 & 80.1 & 72.2 & 50.2 & 78.7 & 65.3 \\
ETN       & 59.2  & 77.0 & 79.5 & 62.9 & 65.7 & 75.0 & 68.3 & 55.4 & 84.4 & 75.7 & 57.7 & 84.5 & 70.5 \\
SAFN      & 58.9  & 76.3 & 81.4 & 70.4 & 73.0 & 77.8 & 72.4 & 55.3 & 80.4 & 75.8 & 60.4 & 79.9 & 71.8 \\
SHOT      & \textbf{64.3}  & 84.0 & \textbf{91.8} & 73.5 & \textbf{77.1} & 84.3 & 78.5 & 64.1 & 89.0 & 80.9 & 63.7 & 86.1 & 78.1 \\
SHOT+IDSP & \textbf{64.3}  & \textbf{85.7} & 91.7 & \textbf{80.9} & \textbf{77.1} & \textbf{88.5} & \textbf{80.5} & \textbf{64.8} & \textbf{89.8} & \textbf{82.3} & \textbf{64.8} & \textbf{87.5} & \textbf{79.9}\\
\hline\hline
\end{tabular}
        \label{tabs3}
\end{table*}

\begin{table*}[]
\centering
\caption{Classification accuracies (\%) on Office-Home dataset for open-set DA (ResNet-50).}
\begin{tabular}{cccccccccccccc}
\hline\hline
Methods   & Ar$\rightarrow$Cl & Ar$\rightarrow$Pr & Ar$\rightarrow$Rw & Cl$\rightarrow$Ar & Cl$\rightarrow$Pr & Cl$\rightarrow$Rw & Pr$\rightarrow$Ar & Pr$\rightarrow$Cl & Pr$\rightarrow$Rw & Rw$\rightarrow$Ar & Rw$\rightarrow$Cl &      Rw$\rightarrow$Pr & AVE\\\hline\hline
ResNet    & 53.4  & 52.7 & 51.9 & 69.3 & 61.8 & 74.1 & 61.4 & 64.0 & 70.0 & 78.7 & 71.0 & 74.0 & 65.3 \\
ATI-$\lambda$     & 55.2  & 52.6 & 53.5 & 69.1 & 63.5 & 74.1 & 61.7 & 64.5 & 70.7 & 79.2 & 72.9 & 75.8 & 66.1 \\
OSBP      & 56.7  & 51.5 & 49.2 & 67.5 & 65.5 & 74.0 & 62.5 & \textbf{64.8} & 69.3 & 80.6 & 74.7 & 71.5 & 65.7 \\
OpenMax   & 56.5  & 52.9 & 53.7 & 69.1 & 64.8 & 74.5 & \textbf{64.1} & 64.0 & 71.2 & 80.3 & 73.0 & 76.9 & 66.7 \\
STA       & 58.1  & 53.1 & 54.4 & \textbf{71.6} & 69.3 & \textbf{81.9} & 63.4 & 65.2 & 74.9 & \textbf{85.0} & \textbf{75.9} & 80.8 & 69.5 \\
SHOT      & \textbf{62.5}  & 81.2 & 83.9 & 60.1 & 74.3 & 80.7 & 63.3 & 57.0 & 82.6 & 70.2 & 64.7 & 82.4 & 71.9 \\
SHOT+IDSP & \textbf{62.5}  & \textbf{83.8} & \textbf{87.8} & 63.8 & \textbf{75.8} & 81.2 & 63.8 & 58.2 & \textbf{83.9} & 73.4 & 68.3 & \textbf{84.0} & \textbf{73.8}\\
\hline\hline
\end{tabular}
        \label{tabs4}
\end{table*}

\subsection{Joint Work with Deep Adaptation Model}
\textcolor{black}{To validate the flexibility of the proposed IDSP, we extend the current DA work (i.e., SHOT \cite{liang2020does}) as a toy by incorporating the IDSP into the corresponding loss function (namely SHOT+IDSP). Specifically, a memory bank is incorporated to store the neighbor information. The implementation is available on GitHub \footnote{\href{https://github.com/Cavin-Lee/DeeP_IDSP}{https://github.com/Cavin-Lee/DeeP\_IDSP}.}. Moreover, we conduct experiments on multiple DA scenarios including closed-set DA, partial-set DA and open-set DA. The benchmark dataset is \emph{Office-Home}}.

\textcolor{black}{Specifically, for the closed-set DA, we adopt DANN \cite{ganin2015unsupervised}, Joint Adaptation Networks (JAN) \cite{long2017deep}, Conditional Adversarial Networks (CDAN) \cite{long2018conditional}, Adversarial Branch Architecture Search (ABAS) \cite{robbiano2022adversarial}, Transferable Attention for Domain Adaptation (TADA) \cite{wang2019transferable},  Aleatoric Certainty based Attention(CADA-A) \cite{kurmi2019attending}, Domain Conditioned Adaptation Network (DCAN) \cite{li2020domain} and SHOT \cite{liang2020does} as the comparison methods. The results are given in Table \ref{tabs2}.}

\textcolor{black}{For the partial-set DA, we adopt Importance Weighted Adversarial Network (IWAN) \cite{zhang2018importance}, Selective Adversarial Network (SAN) \cite{cao2018partialCVPR}, Example Transfer Network (ETN) \cite{cao2019learning}, Stepwise Adaptive Feature Norm (SAFN) \cite{xu2019larger} and SHOT \cite{liang2020does} as the comparison methods.The results are given in Table \ref{tabs3}}

\textcolor{black}{For the open-set DA, we adopt Assign and Transform Iteratively (ATI-$\lambda$) \cite{panareda2017open},Open Set domain adaptation by BackPropagation (OSBP) \cite{saito2018open}, OpenMax \cite{bendale2016towards},Separate to Adapt (STA) \cite{liu2019separate} and SHOT \cite{liang2020does} as the comparison methods.The results are given in Table \ref{tabs4}.}

\textcolor{black}{The experimental setting is referring to SHOT. As we can observe that, in terms of average accuracy, the proposed IDSP helps SHOT achieve better performance: +2\% on closed-set UDA task, +1.8\% on PDA task, and +1.9\% on open-set DA task. The results illustrate that the IDSP can enhance the adaptation performance of both deep and shallow model, which confirms the effectiveness of model smoothness.}

\section{Graph Construction for Structure Preserving}
 In order to demonstrate the effectiveness of the proposed Intra Domain Structure Preserving, we perform an ablation study by learning a classifier with different structure preserving constraints. Specifically, we use no structure preserving (nP, i.e., $\gamma$ = 0), manifold structure preserving of source and target domain (CST), intra structure of both source and target domain (ST) and intra structure preserving of target domain (T). The pair-wise affinity matrices $\mathbf{G}^{(T)}$, $\mathbf{G}^{(ST)}$ and $\mathbf{G}^{(CST)}$ of the graph can be formulated as follows:

\begin{equation}
  \begin{aligned}
  \label{eq7}
G^{(T)}_{i j} & = \left\{\begin{array}{ll}
\hat{G}_{i j}, & \mathbf{x}_{i} \text { and } \mathbf{x}_{j} \in \mathcal{D}_{t}  \\
0, & \text { otherwise }
\end{array}\right.
\end{aligned}.
\end{equation}
 The pair-wise affinity matrix $\hat{G}^{(ST)}$ of the graph can be formulated as follows:
\begin{equation}
  \begin{aligned}
  \label{eqere}
G^{(ST)}_{i j} & = \left\{\begin{array}{ll}
\hat{G}_{i j}, & \left(\mathbf{x}_{i}, \mathbf{x}_{j}\right) \in \mathcal{D}_{t} \text{or}  \left(\mathbf{x}_{i}, \mathbf{x}_{j}\right) \in \mathcal{D}_{s}\\
0, & \text { otherwise }
\end{array}\right.
\end{aligned}.
\end{equation}
 The pair-wise affinity matrix $\hat{G}^{(CST)}$ of the graph can be formulated as follows:
\begin{equation}
  G^{(CST)}_{i j}  = \hat{G}_{i j}.
\end{equation}
\bibliographystyle{ieeetr}
\bibliography{mybibfile}